# Reasoning with Topological and Directional Spatial Information

Sanjiang Li[a*] and Anthony G Cohn[b]

[a] Centre for Quantum Computation and Intelligent Systems,
Faculty of Engineering and Information Technology,
University of Technology, Sydney, Australia

[b] School of Computing, University of Leeds, Leeds, LS2 9JT, UK

lisj@it.uts.edu.au (S. Li)   agc@comp.leeds.ac.uk (A.G. Cohn)

October 12, 2018

## Abstract

Current research on qualitative spatial representation and reasoning mainly focuses on one single aspect of space. In real world applications, however, multiple spatial aspects are often involved simultaneously.

This paper investigates problems arising in reasoning with combined topological and directional information. We use the RCC8 algebra and the Rectangle Algebra (RA) for expressing topological and directional information respectively. We give examples to show that the bipath-consistency algorithm BIPATH-CONSISTENCY is incomplete for solving even basic RCC8 and RA constraints. If topological constraints are taken from some maximal tractable subclasses of RCC8, and directional constraints are taken from a subalgebra, termed DIR49, of RA, then we show that BIPATH-CONSISTENCY is able to *separate* topological constraints from directional ones. This means, given a set of hybrid topological and directional constraints from the above subclasses of RCC8 and RA, we can transfer the joint satisfaction problem in polynomial time to two independent satisfaction problems in RCC8 and RA. For general RA constraints, we give a method to compute solutions that satisfy all topological constraints and approximately satisfy each RA constraint to any prescribed precision.

## 1 Introduction

Originating from Allen's work on temporal interval relations [1], the qualitative approach to temporal as well as spatial information is popular in Artificial Intelligence and related research fields. This is mainly because precise numerical

---

[*]Corresponding Author

information is often unavailable or not necessary in many real world applications [4, 5].

Typically, the qualitative approach represents temporal and spatial information by introducing a (binary) relation model on the universe of temporal or spatial entities, which contains a finite set of binary relations defined on the universe. Finding a proper relation model, or a *qualitative calculus*, is the key to the success of the qualitative approach to temporal and spatial reasoning. This is partially justified by the great success of Allen's Interval Algebra (IA), which is the principal formalism of qualitative temporal reasoning.

As for spatial reasoning, dozens of spatial relation models have been developed in the past twenty years. Since relations in the same model are ideally homogenous, most spatial calculi focus on one single aspect of space, e.g. topology, direction, distance, or position. When representing spatial direction, distance and position, it is convenient to approximate spatial entities by points. But this is inappropriate as far as spatial topological information is concerned: topology concerns sets of points, i.e. regions.

Topological relations are invariant under homeomorphism such as scale, rotation, and translation. It is widely acknowledged that topological relations are of crucial importance, and the slogan is *"topology matters, metric refines* [9]." An influential formalism for topological relations is the Region Connection Calculus (RCC) [31]. RCC represents spatial entities as arbitrary plane[1] regions, which may have holes or have multiple connected components. Based on one primitive binary connectedness relation, a set of eight jointly exhaustive and pairwise disjoint (JEPD) relations can be defined in RCC. The Boolean algebra generated by this set is known as the RCC8 algebra. A similar formalism is the 9-Intersection Method (9IM) of Egenhofer [8], where the same eight relations are defined on simple plane regions (regions homeomorphic to a closed disk). This relation model, called the Egenhofer model in [23], is widely used in geographical information science.

The RCC8 algebra and the Egenhofer model only represent the topological information between spatial objects. But in many practical applications and particularly in natural language expressions, topological relations are used together with other kinds of spatial relations. For example, when describing the location of Titisee, a famous tourist sight in Germany, we might say "Titisee is *in* the Black Forest and is *east* of the town of Freiburg." In order to provide a more expressive formalism for spatial information, it is necessary to combine different kinds of spatial information.

The major obstacle to the combination is how to reason with combined information efficiently. An important reasoning problem is the *joint satisfaction problem* (JSP). Suppose $\mathfrak{A}$ and $\mathfrak{B}$ are two relation models over the same universe. Given two networks of constraints over $\mathfrak{A}$ and $\mathfrak{B}$, respectively, decide if there exists a common solution to both networks.

In order to solve the joint satisfaction problem over $\mathfrak{A}$ and $\mathfrak{B}$, one natural way is to define a hybrid relation model $\mathfrak{C}$ which is the smallest Boolean algebra containing both $\mathfrak{A}$ and $\mathfrak{B}$ and to reason with $\mathfrak{C}$ by the usual composition-based reasoning techniques. Although the (weak) composition table of the hybrid model can be established as usual, composition-based reasoning is often incomplete for deciding if a constraint network is satisfiable. Moreover, it will be difficult to make use of the techniques already developed for the two component models.

---

[1] RCC can in fact be used to reason about regions of any dimension, providing they are all of the same dimension, but here we focus on 2D regions.

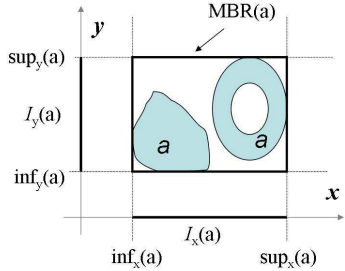

Figure 1: Illustrations of a bounded region $a$ and its minimum bounding rectangle $\mathsf{MBR}(a)$, where $a$ contains a hole and has two connected components.

Instead of developing a new hybrid calculus, this work deals with the joint satisfaction problem directly. We concern ourselves with the combination of topological and directional relations, since these are the two most important kinds of spatial relations.

We represent *extended* spatial objects as bounded plane regions and adopt the RCC8 Algebra to model topological relations. To represent directional information, we need to define a direction relation model. One natural requirement for such a relation model is that it should support definitions of cardinal directions over extended objects. Unlike topological relations such as *partially overlap* and *non-tangentially proper part*, which have unambiguous semantics, researchers have no agreement on the definitions of cardinal directions such as *west*, *east*, *north*, and *south*. Several different interpretations of cardinal directions over extended objects have been given in the literature [14, 30, 39, 41].

This paper, following Sistla, Yu, and Haddad [39], takes the projection-based definition of cardinal directions. For an extended object $a$, we project $a$ to the two predefined orthogonal basis in the real plane (see Figure 1), and write $I_x(a)$ and $I_y(a)$ for the smallest convex intervals which contain the projections of $a$ on the $x$- and $y$-axis, respectively. For two extended objects $b$ and $c$, we say $b$ is *west* of object $c$ if $I_x(b)$ is *before* $I_x(c)$, i.e. the right endpoint of $I_x(b)$ is smaller than the left endpoint of $I_x(c)$. The other cardinal directions are defined in a similar way.

A more expressive representation of direction relations can be obtained by using an extension of the Rectangle Algebra (RA) [15], which is the two dimensional generalization of IA. For an extended object $a$, we write $\mathsf{MBR}(a) =$ $I_x(a) \times I_y(a)$ for the *minimum bounding rectangle* of $a$ (see Figure 1). The extended rectangle relation between $b, c$ is defined by the IA relation $\lambda_x$ between $I_x(b)$ and $I_x(c)$ and the IA relation $\lambda_y$ between $I_y(b)$ and $I_y(c)$. For convenience, we write $\lambda_x \otimes \lambda_y$ for the extended rectangle relation between $b$ and $c$. In what follows, we call this model of relations on bounded plane regions the *Extended Rectangle Algebra* (ERA).

We now have two relation models — RCC8 and ERA — defined on the same universe of bounded plane regions. The next step is to find efficient and complete methods for solving the joint satisfaction problem (JSP). Recall that the two independent satisfaction problems over RCC8 and ERA are NP-complete and large tractable subclasses of RCC8 and ERA have been found [34, 2]. The JSP over RCC8 and ERA is more difficult than the two independent satisfaction problems. This is because different aspects may interact with each other, and two independently satisfiable networks may be jointly unsatisfiable. For example, suppose $a, b, c, d$ are four spatial objects, and the only topological information we know is that $a$ partially overlaps $c$, and $b$ partially overlaps $d$. Somehow, an outdated map also suggests that $a$ is west of $b$, and $c$ is east of $d$. The two topological (directional) constraints are apparently satisfiable. But when combined the four constraints are unsatisfiable.

The JSP over RCC8 and ERA has been investigated to some extent by several researchers. Sharma [37] discussed the problem where at most three variables are involved. Sistla et al. [39, 38] established a complete decision method for the small set of relations that consists of the four cardinal directions and part-whole relations *inside*, *outside*, and *overlaps*.[2] Therefore, more work is needed to solve the JSP over RCC8 and ERA.

We introduce the notions of bi-closure and bipath-consistency to process hybrid spatial constraints locally. These two notions are similar to the well-known arc- and path-consistency in constraint solving (cf. [6]). Bi-closure concerns the satisfiablity of constraints defined on any two variables, while bipath-consistency concerns the satisfiablity of constraints defined on any three variables. Applying the bipath-consistency algorithm BIPATH-CONSISTENCY introduced in [12], we can transfer a joint network of RCC8 and ERA constraints in cubic time to another bipath-consistent (bi-closed, resp.) joint network that has the same solutions.

Ideally, we would hope BIPATH-CONSISTENCY provides a complete solving technique for the whole RCC8 Algebra and ERA. Examples show, however, this is not true. In the absence of such a result, we turn to finding large subclasses of RCC8 and ERA. In this paper, we introduce a subalgebra —DIR49— of ERA, which contains forty-nine basic relations and supports the definition of cardinal direction relations. DIR49 is the two dimensional counterpart of the interval algebra $\mathsf{IA}_7$, proposed in [13], where each basic relation of $\mathsf{IA}_7$ is the union of several 'similar' basic IA relations.

We then show that BIPATH-CONSISTENCY can be used to solve RCC8 and DIR constraints simultaneously. Recall that $\widehat{\mathcal{H}}_8$ is one of the three maximal tractable subclass of RCC8 that contains all the basic relations [33]. Let $\mathcal{N}_{top}$ be an RCC8 network over $\widehat{\mathcal{H}}_8$, and let $\mathcal{N}_{dir}$ be an RA network over DIR49. Suppose $(\mathcal{N}'_{top}, \mathcal{N}'_{dir})$ is a bipath-consistent network that has the same solutions with $(\mathcal{N}_{top}, \mathcal{N}_{dir})$. Then we show (Theorem 6.4) $(\mathcal{N}_{top}, \mathcal{N}_{dir})$ is satisfiable if and only if both the RCC8 network $\mathcal{N}'_{top}$ and the RA network $\mathcal{N}'_{dir}$) are independently satisfiable. The JSP of an arbitrary RCC8 network and a DIR49 network can then be determined by backtracking RCC8 constraints over $\widehat{\mathcal{H}}_8$. This means that reasoning with DIR49 and RCC8 is an NP problem.

---
[2]These correspond to the RCC8 relations *part of* (**P**), *disconnected from* (**DC**), and *partially overlaps* (**PO**).

The general JSP over RCC8 and ERA can also be tackled in an approximate sense. Suppose $V = \{v_i\}_{i=1}^n$ is a set of variables, and suppose $\mathcal{N}_{top} = \{v_i \theta_{ij} v_j\}_{i,j=1}^n$ and $\mathcal{N}_{dir} = \{v_i \delta_{ij} v_j\}_{i,j=1}^n$ are two networks of constraints over RCC8 and ERA, respectively. If $\mathcal{N}_{top} \cup \mathcal{N}_{dir}$ is satisfiable, then we can find a solution $\{a_i\}_{i=1}^n$ of $\mathcal{N}_{top}$ that *almost* satisfies each constraint $\delta_{ij}$ in $\mathcal{N}_{dir}$ with any prescribed precision. This means, a slight change (e.g. by translating or enlarging $a_i$) may make $(a_i, a_j)$ an instance of $\delta_{ij}$ for any $i, j$.

The remainder of this paper proceeds as follows. Section 2 introduces basic notions and well-known examples of qualitative calculi, including IA, RCC8, RA etc. Section 3 extends the universe of Rectangle Algebra from rectangles to general bounded regions. The resulted calculus is termed ERA. We also define the subalgebra DIR49 of ERA. Section 4 proposes the combination problem of two qualitative calculi. The notions of bi-closure and bipath-consistency are introduced in this section. In this section we also show by examples that the bipath-consistency algorithm is not complete for determining the joint satisfaction problem over RCC8 and ERA. We then describe how to compute the bi-closure for a pair of RCC8 and ERA constraints in Section 5, and prove how BIPATH-CONSISTENCY separate $\widehat{\mathcal{H}}_8$ from DIR49 in Section 6. Section 7 exploits this separation theorem to cope with the general JSP over RCC8 and ERA. Section 8 discusses the related work and Section 9 concludes the paper.

This work greatly extends an earlier paper reported at IJCAI-07 [21], where separation theorems were obtained for a quite small subalgebra of DIR49 and all maximal tractable subclasses of RCC8.

## 2 Qualitative Calculi

The establishment of a proper qualitative calculus is the key to the success of the qualitative approach to temporal and spatial reasoning. This section introduces basic notions and important examples of qualitative calculi (see also [25]).

### 2.1 Basic Notions

Let $D$ be a universe of temporal or spatial or spatial-temporal entities. We use small Greek symbols for representing relations on $D$. For a relation $\alpha$ on $D$ and two elements $x, y$ in $D$, we write $(x, y) \in \alpha$ or $x \alpha y$ to indicate that $(x, y)$ is an instance of $\alpha$. For two relations $\alpha, \beta$ on $D$, we define the complement of $\alpha$, the intersection, and the union of $\alpha$ and $\beta$ as follows.

$$\begin{aligned} -\alpha &= \{(x,y) \in D \times D : (x,y) \notin \alpha\} \\ \alpha \cap \beta &= \{(x,y) \in D \times D : (x,y) \in \alpha \text{ and } (x,y) \in \beta\} \\ \alpha \cup \beta &= \{(x,y) \in D \times D : (x,y) \in \alpha \text{ or } (x,y) \in \beta\}. \end{aligned}$$

We write $\mathbf{Rel}(D)$ for the set of binary relations on $D$. Clearly, the 6-tuple $(\mathbf{Rel}(D); -, \cap, \cup, \varnothing, D \times D)$ is a Boolean algebra, where $\varnothing$ and $D \times D$ are, respectively, the empty relation and the universal relation on $D$.

A finite set $\mathcal{B}$ of nonempty relations on $D$ is *jointly exhaustive and pairwise disjoint* (JEPD) if any two entities in $D$ are related by one and only one relation in $\mathcal{B}$. We write $\langle\langle\mathcal{B}\rangle\rangle$ for the subalgebra of $\mathbf{Rel}(D)$ generated by $\mathcal{B}$, i.e. the smallest subalgebra of the Boolean algebra $\mathbf{Rel}(D)$ which contains $\mathcal{B}$. Clearly, relations in $\mathcal{B}$ are atoms in the Boolean algebra $\langle\langle\mathcal{B}\rangle\rangle$. We call $\langle\langle\mathcal{B}\rangle\rangle$ a *qualitative calculus* on $D$, and call relations in $\mathcal{B}$ *basic* relations of the calculus.

We write $id_D$ for the identity relation on $D$. For two relations $\alpha, \beta$ on $D$, we define the converse of $\alpha$ and the composition of $\alpha$ and $\beta$ as follows.

$$\begin{aligned} \alpha^\smile &= \{(y,x) \in D \times D : (x,y) \in \alpha\} \\ \alpha \circ \beta &= \{(x,y) \in D \times D : (\exists z \in D) [(x,z) \in \alpha \text{ and } (z,y) \in \beta]\}. \end{aligned}$$

*Remark* 2.1. Our definition of a qualitative calculus is more general than the one given by Ligozat and Renz [25], where the set $\mathcal{B}$ is required to be closed under converse and contain the identity relation $id_D$. There are several relation models that do not satisfy these conditions. One example is the cardinal direction calculus (CDC) [14], another is the Extended Rectangle Algebra (ERA) (to be introduced in Section 3.1).

Note that the composition of two relations in $\langle\langle\mathcal{B}\rangle\rangle$ is not necessarily in $\langle\langle\mathcal{B}\rangle\rangle$. For $\alpha, \beta \in \langle\langle\mathcal{B}\rangle\rangle$, the *weak composition* [7, 24] of $\alpha$ and $\beta$, written as $\alpha \circ_w \beta$, is defined to be the smallest relation in $\langle\langle\mathcal{B}\rangle\rangle$ which contains $\alpha \circ \beta$. We say a qualitative calculus $\langle\langle\mathcal{B}\rangle\rangle$ is *closed under composition* if the composition of any two relations in $\langle\langle\mathcal{B}\rangle\rangle$ is still a relation in $\langle\langle\mathcal{B}\rangle\rangle$. This is equivalent to saying that the weak composition operation is the same as the composition operation.

An important reasoning problem in a qualitative calculus $\langle\langle\mathcal{B}\rangle\rangle$ is the satisfaction problem. Let $\mathcal{A}$ be a subset of $\langle\langle\mathcal{B}\rangle\rangle$. A constraint over $\mathcal{A}$ has the form $(x\gamma y)$ with $\gamma \in \mathcal{A}$. For a set of variables $V = \{v_i\}_{i=1}^n$, and a set of constraints $\mathcal{N}$ involving variables in $V$, we say $\mathcal{N}$ is a *constraint network* if for each pair $(i,j)$ there exists a unique constraint $(x_i \gamma x_j)$ in $\mathcal{N}$. A network $\mathcal{N}$ is said to be over $\mathcal{A}$ if each constraint in $\mathcal{N}$ is over $\mathcal{A}$. We say a constraint network $\mathcal{N} = \{v_i \gamma_{ij} v_j\}_{i,j=1}^n$ is *satisfiable* (or *consistent*) if there is an instantiation $\{a_i\}_{i=1}^n$ in $D$ such that $(a_i, a_j) \in \gamma_{ij}$ holds for all $1 \leq i, j \leq n$. In this case, we call $\{a_i\}_{i=1}^n$ a solution of $\mathcal{N}$. The satisfaction problem over $\mathcal{A}$ is the decision problem of the satisfiability of constraint networks over $\mathcal{A}$.

For two constraint networks $\mathcal{N} = \{v_i \gamma_{ij} v_j\}_{i,j=1}^n$ and $\mathcal{N}' = \{v_i \gamma'_{ij} v_j\}_{i,j=1}^n$ over $\langle\langle\mathcal{B}\rangle\rangle$, we say $\mathcal{N}$ and $\mathcal{N}'$ are *equivalent* if they have the same set of solutions, and say $\mathcal{N}'$ *refines* $\mathcal{N}$ if each constraint $\gamma'_{ij}$ is contained in $\gamma_{ij}$. If $\mathcal{N}'$ refines $\mathcal{N}$ and each $\gamma'_{ij}$ is a basic relation in $\mathcal{B}$, then we call $\mathcal{N}'$ a *scenario* of $\mathcal{N}$.

Table 1: The set of basic interval relations $\mathcal{B}_{int}$, where $x = [x^-, x^+], y = [y^-, y^+]$ are two intervals.

| Relation | Symb. | Conv. | Meaning |
|---|---|---|---|
| before | b | bi | $x^+ < y^-$ |
| meets | m | mi | $x^+ = y^-$ |
| overlaps | o | oi | $x^- < y^- < x^+ < y^+$ |
| starts | s | si | $x^- = y^- < x^+ < y^+$ |
| during | d | di | $x^- < y^- < y^+ < x^+$ |
| finishes | f | fi | $y^- < x^- < x^+ = y^+$ |
| equals | eq | eq | $x^- = y^- < x^+ = y^+$ |

The consistency of a network can be approximated by using a cubic path-consistency algorithm (PCA). A network $\mathcal{N} = \{v_i \gamma_{ij} v_j\}_{i,j=1}^n$ is *path-consistent* if every subnetwork containing at most three variables is consistent. The essence of a PCA is to apply the following updating rule for all $i, j, k$ until the network is stable [1, 22].

$$\gamma_{ij} \leftarrow \gamma_{ij} \cap \gamma_{ik} \circ_w \gamma_{kj} \qquad (1)$$

If the empty relation occurs during the process, then the network is inconsistent, otherwise the resulting network is path-consistent.

## 2.2 Interval Algebra

The Interval Algebra (IA) [1] is generated by a set $\mathcal{B}_{int}$ of 13 basic relations between time intervals (see Table 1). We call relations in IA interval relations. Two basic interval relations in $\mathcal{B}_{int}$ are *conceptual neighbors* [10] if they can be directly transformed into one another by continuous deformation. Different kinds of deformations may give rise to different conceptual neighborhood graphs (CNGs). Figure 2 shows the CNG induced by fixing three of the four endpoints of two events while moving the fourth.

A set of basic interval relations is called a *conceptual neighborhood* [10] if its elements are path-connected in the CNG. By Figure 2, we know m is a neighbor of o, and s and f are two neighbors of d. As a consequence, {m, o} and {s, d, f} are two conceptual neighborhoods.

Each neighborhood corresponds to an interval relation. The following non-basic interval relations are all induced by some neighborhoods:

$$\begin{aligned}
(\text{mo}) &= \text{m} \cup \text{o} \\
(\text{sfd}) &= \text{s} \cup \text{f} \cup \text{d} \\
(\text{sfdeq}) &= \text{s} \cup \text{f} \cup \text{d} \cup \text{eq} \\
\pitchfork &= \text{m} \cup \text{o} \cup \text{s} \cup \text{f} \cup \text{d} \cup \text{eq} \cup \text{di} \cup \text{fi} \cup \text{si} \cup \text{oi} \cup \text{mi}.
\end{aligned}$$

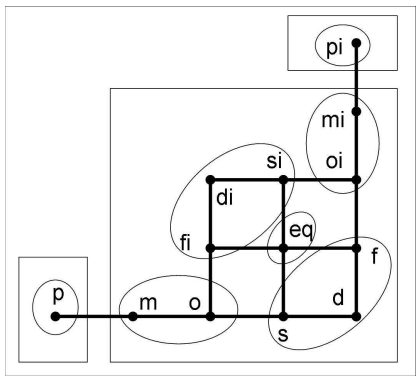

Figure 2: The conceptual neighborhood graph of Interval Algebra [10], where ellipses (boxes, resp.) represent basic relations in $\mathsf{IA}_7$ ($\mathsf{IA}_3$, resp.).

These non-basic relations, as well as their converses, are frequently used in this paper. Let

$$\mathcal{B}^3_{int} = \{\mathsf{b}, \mathsf{m}, \mathsf{bi}\} \tag{2}$$

$$\mathcal{B}^7_{int} = \{\mathsf{b}, (\mathsf{mo}), (\mathsf{sfd}), \mathsf{eq}, (\mathsf{sfd})^\sim, (\mathsf{mo})^\sim, \mathsf{bi}\} \tag{3}$$

It is clear that both $\mathcal{B}^3_{int}$ and $\mathcal{B}^7_{int}$ are JEPD sets of interval relations. Moreover, relations in $\mathcal{B}_3$ and $\mathcal{B}_7$ are all conceptual neighborhoods in the sense of Freksa [10]. Write $\mathsf{IA}_3$ and $\mathsf{IA}_7$ for the Boolean algebras generated by these two sets, respectively. These two algebras, first introduced by Golumbic and Shamir [13], provide two coarser versions of IA. Moreover, they also proved that $\mathsf{IA}_3$ and $\mathsf{IA}_7$ are intractable, and

$$\mathcal{H}_3 = \{\mathsf{b}, \mathsf{m}, \mathsf{bi}, \mathsf{b} \cup \mathsf{m}, \mathsf{m} \cup \mathsf{bi}, \top\} \tag{4}$$

is a maximal tractable subclass of $\mathsf{IA}_3$ [13], where $\top$ is the universal relation.

Nebel and Bürckert [28] identified a maximal tractable subclass $\mathcal{H}$ of IA, called the *ORD-Horn* subclass, and showed that applying PCA is sufficient for the satisfaction problem over $\mathcal{H}$. It is straightforward to show that $\mathcal{H}_3$ is the intersection of $\mathcal{H}$ and $\mathsf{IA}_3$. Let $\mathcal{H}_7 \equiv \mathcal{H} \cap \mathsf{IA}_7$. As a subset of $\mathcal{H}$, $\mathcal{H}_7$ is also a tractable subclass of $\mathsf{IA}_7$.

*Remark* 2.2. While IA is closed under composition, the two subalgebras $\mathsf{IA}_3$ and $\mathsf{IA}_7$ are not. Therefore, they are not coarser calculi of IA in the sense of [36]. For our purposes this is not a problem. For a subalgebra like $\mathsf{IA}_3$ or $\mathsf{IA}_7$, the most important thing is that it provides an abstraction for relations in IA at a reasonable granularity.

As for the reasoning aspect, the (weak) composition-based reasoning techniques are incomplete for these subalgebras. But other efficient and complete reasoning techniques exist. For example, Golumbic and Shamir [13] proposed a graph-theoretic approach for solving the constraint satisfaction problem of $\mathsf{IA}_3$, which determines the satisfiability of a constraint network over $\mathcal{H}_3$ in polynomial time.

Moreover, complete reasoning techniques for IA, e.g. the path-consistency algorithm, can be applied to solving the satisfaction problem of any subalgebra of IA. This clearly provides a complete reasoning method for the subalgebra. But when restricted to the subalgebra, the reasoning method may be not efficient even for solving constraint problems that only involve basic relations in the subalgebra. This is because basic relations of the subalgebra may be outside the ORD-Horn subclass $\mathcal{H}$ of IA. But for $\mathsf{IA}_3$ and $\mathsf{IA}_7$, we know $\mathcal{B}_3$ and $\mathcal{B}_7$ are contained in $\mathcal{H}$. Therefore, the path-consistency algorithm developed for IA can be applied to solving reasoning problems over $\mathcal{H}_3$ and $\mathcal{H}_7$ efficiently.

### 2.3 RCC8 Algebra

A *plane region* (or *a region*) is a nonempty regular closed subset of the real plane. A region is *bounded* if it is contained in a disk. In this paper, we only consider bounded regions. Let $U$ be the set of bounded regions. The relations defined in Table 2 and the converses of **TPP** and **NTPP** form a JEPD set of relations on $U$. These are the RCC8 basic relations. Write $\mathcal{B}_{top}$ for this set. The RCC8 Algebra [31] is the subalgebra of $\mathbf{Rel}(U)$ generated by $\mathcal{B}_{top}$. We write **P** and **PP**, resp., for $\mathbf{TPP} \cup \mathbf{NTPP} \cup \mathbf{EQ}$ and $\mathbf{TPP} \cup \mathbf{NTPP}$.

Table 2: The set of RCC8 basic relations $\mathcal{B}_{top}$, where $a, b$ are two bounded regions and $a°$ and $b°$ are, resp., their interiors.

| Relation | Symb. | Meaning |
|---|---|---|
| equals | **EQ** | $a = b$ |
| disconnected | **DC** | $a \cap b = \varnothing$ |
| externally connected | **EC** | $a \cap b \neq \varnothing \ \wedge \ a° \cap b° = \varnothing$ |
| partially overlap | **PO** | $a° \cap b° \neq \varnothing \ \wedge \ a \not\subseteq b \ \wedge \ a \not\supseteq b$ |
| tangential proper part | **TPP** | $a \subset b \ \wedge \ a \not\subset b°$ |
| non-tangential proper part | **NTPP** | $a \subset b°$ |

The satisfaction problem over the whole RCC8 Algebra is NP-complete, but three maximal tractable subclasses of RCC8 have been found [33]. These subclasses, denoted by $\widehat{\mathcal{H}}_8, \mathcal{C}_8, \mathcal{Q}_8$, are the only maximal tractable subclasses which contain all basic relations. For these subclasses, applying PCA is sufficient for deciding the satisfiability of a network. Moreover, for a path-consistent network over one of the three maximal tractable subclasses, we can find a satisfiable scenario in $O(n^2)$ time [33].

### 2.4 Qualitative Size Calculus

A qualitative size calculus [12] can be defined on the set $U$ of bounded regions. For two bounded regions $a, b$, the size of $a$ is said to be smaller than that of $b$, denoted by $a <_s b$, if the area of $a$ is smaller than that of $b$. The definitions of $a =_s b$ and $a >_s b$ are similar. Write QS for the qualitative calculus on $U$ generated by the JEPD set of relations $\{<_s, =_s, >_s\}$. It is clear that QS is another representation for the well-known Point Algebra [29].

## 2.5 Rectangle Algebra

The Rectangle Algebra (RA) [15, 2] is a qualitative calculus defined on the set of all rectangles in the plane, where we assume that the two sides of a rectangle are parallel to the axes of some predefined orthogonal basis in the Euclidean plane.

For a rectangle $r$, write $I_x(r)$ and $I_y(r)$ as, resp., the $x$- and $y$-projection of $r$. The basic rectangle relation between two rectangles $r_1, r_2$ is defined by the basic IA relation between $I_x(r_1)$ and $I_x(r_2)$ and that between $I_y(r_1)$ and $I_y(r_2)$. More precisely, if $(I_x(r_1), I_x(r_2)) \in \alpha$ and $(I_y(r_1), I_y(r_2)) \in \beta$, then we write $\alpha \otimes \beta$ for the basic rectangle relation between $r_1$ and $r_2$. In other words, for any basic IA relations $\alpha$, $\beta$,

$$(r_1, r_2) \in \alpha \otimes \beta \Leftrightarrow (I_x(r_1), I_x(r_2)) \in \alpha \ \& \ (I_y(r_1), I_y(r_2)) \in \beta. \tag{5}$$

Write $\mathcal{B}_{rec}$ for the set of these rectangle relations, i.e.

$$\mathcal{B}_{rec} = \{\alpha \otimes \beta : \alpha, \beta \in \mathcal{B}_{int}\} \tag{6}$$

RA is then the qualitative calculus generated by $\mathcal{B}_{rec}$ on the set of rectangles.

*Remark* 2.3. If $\mathcal{S}$ is a tractable subclass of IA, then $\mathcal{S} \otimes \mathcal{S} = \{\alpha \otimes \beta : \alpha, \beta \in \mathcal{S}\}$ is also tractable in RA. This is because, a basic RA network $\mathcal{N} = \{v_i \alpha_{ij} \otimes \beta_{ij} v_j\}_{i,j=1}^n$ ($\alpha_{ij}, \beta_{ij} \in \mathcal{B}_{int}$) is satisfiable iff both of its component IA networks $\mathcal{N}_x = \{v_i \alpha_{ij} v_j\}_{i,j=1}^n$ and $\mathcal{N}_y = \{v_i \beta_{ij} v_j\}_{i,j=1}^n$ are satisfiable. A tractable subclass of RA larger than $\mathcal{H} \otimes \mathcal{H}$ is obtained in [2], where $\mathcal{H}$ is the ORD-Horn subclass of IA.

In the next section, we will introduce several qualitative direction calculi.

## 3 Cardinal Direction Calculus

RA can be adapted for representing directional information. To this end, we first extend the universe of RA from the set of rectangles to the set of bounded regions, and then formalize the four cardinal directions, and lastly introduce two coarser direction calculi.

## 3.1 The Extended Rectangle Algebra ERA

We begin with the notion of a *minimum bounding rectangle* (MBR). For a bounded region $a$, define (see Figure 1)

$$\sup_x(a) = \sup\{x \in \mathbb{R} : (\exists y)(x,y) \in a\}, \tag{7}$$

$$\inf_x(a) = \inf\{x \in \mathbb{R} : (\exists y)(x,y) \in a\}, \tag{8}$$

$$\sup_y(a) = \sup\{y \in \mathbb{R} : (\exists x)(x,y) \in a\}, \tag{9}$$

$$\inf_y(a) = \inf\{y \in \mathbb{R} : (\exists x)(x,y) \in a\}. \tag{10}$$

Write $I_x(a) = [\inf_x(a), \sup_x(a)]$ and $I_y(a) = [\inf_y(a), \sup_y(a)]$ for the $x$- and $y$-projection of $a$. We call $I_x(a) \times I_y(a)$ the *minimum bounding rectangle* (MBR) of $a$, denoted by $\mathsf{MBR}(a)$.

For two bounded regions $a, b$, we define the *extended rectangle relation* between $a, b$ as the rectangle relation between $\mathsf{MBR}(a)$ and $\mathsf{MBR}(b)$. To avoid introducing new notation, we use the same relation symbol, i.e. for a rectangle relation $\alpha$,

$$a\alpha b \Leftrightarrow \mathsf{MBR}(a)\alpha\mathsf{MBR}(b). \tag{11}$$

In this way, we extend the universe of RA from the set of rectangles to $U$, the set of bounded regions. We call this calculus the *Extended Rectangle Algebra*, written ERA.

Clearly, a network $\mathcal{N} = \{v_i \delta_{ij} v_j\}_{i,j=1}^n$ of constraints over ERA could also be interpreted as a constraint network over RA. This will cause no trouble since $\{a_i\}_{i=1}^n$ is a solution to the ERA network $\mathcal{N}$ iff $\{\mathsf{MBR}(a_i)\}_{i=1}^n$ is a solution to the RA network $\mathcal{N}$. Moreover, if $\{r_i\}_{i=1}^n$ is a solution to the RA network $\mathcal{N}$, then it is also a solution to the ERA network $\mathcal{N}$. In this case, we also call $\{r_i\}_{i=1}^n$ a *rectangle solution* of $\mathcal{N}$.

**Lemma 3.1.** *A network $\mathcal{N}$ of ERA constraints is satisfiable if and only if $\mathcal{N}$ is satisfiable as an RA constraint network. In other words, $\mathcal{N}$ has a solution in $U$ if and only if it has a rectangle solution.*

ERA provides a natural representation for directional information among extended regions. In particular, the four cardinal directions can be represented as (non-basic) relations in ERA. To show this, we first formalize the four cardinal directions.

**Definition 3.1.** For two bounded regions $a, b$, if $\sup_x(a) < \inf_x(b)$, then we say $a$ is *west* of $b$ and $b$ is *east* of $a$, written as $a\mathsf{W}b$ and $b\mathsf{E}a$; and if $\sup_y(a) < \inf_y(b)$ then we say $a$ is *south* of $b$ and $b$ is *north* of $a$, written as $a\mathsf{S}b$ and $b\mathsf{N}a$.

Then, take $\mathsf{W}$ as an example (see Figure 3). It is clear that $\mathsf{W}$ is the union of all rectangle relations $\mathsf{b} \otimes \alpha$ with $\alpha \in \mathcal{B}_{int}$, and therefore a relation in ERA. Note that other well-known directional relations such as *northwest* can be defined as the intersection of cardinal directions *north* and *west*.

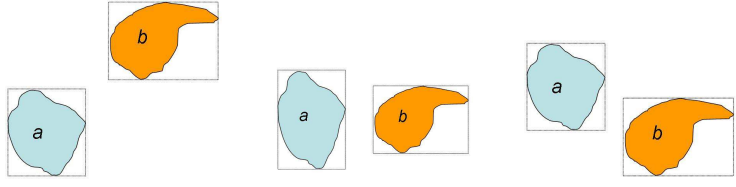

Figure 3: Illustrations of the cardinal direction *West*: $(a,b) \in \mathsf{b} \otimes \mathsf{b}$ (left), $(a,b) \in \mathsf{b} \otimes \mathsf{di}$ (center), $(a,b) \in \mathsf{b} \otimes \mathsf{oi}$ (right).

## 3.2 Two Simpler Direction Calculi: DIR9 and DIR49

Although ERA provides a very expressive formalism for directional relations, it is perhaps too complicated to use in practical applications. In these situations, simplified versions are more desirable. In this subsection, we introduce two coarser calculi of ERA.

Recall that ⋒ stands for the union of all basic interval relations other than b and bi. It is easy to see that the relations in

$$\mathcal{B}_{rec}^{9} = \{\mathsf{b} \otimes \mathsf{b}, \mathsf{b} \otimes \mathsf{⋒}, \mathsf{b} \otimes \mathsf{bi}, \mathsf{⋒} \otimes \mathsf{b}, \mathsf{⋒} \otimes \mathsf{⋒}, \mathsf{⋒} \otimes \mathsf{bi}, \mathsf{bi} \otimes \mathsf{b}, \mathsf{bi} \otimes \mathsf{⋒}, \mathsf{bi} \otimes \mathsf{bi}\} \quad (12)$$

are atoms of the Boolean algebra generated by N,S,W,E. We write DIR9 for this subalgebra of ERA. Although it is very simple, DIR9 is sufficient for expressing directional information in many situations. Moreover, all direction relations which appeared in [38] can be expressed in DIR9.

DIR9 is the two-dimensional counterpart of $\mathsf{IA}_3$ — the subalgebra of IA generated by $\mathcal{B}_{int}^{3} = \{\mathsf{b}, \mathsf{⋒}, \mathsf{bi}\}$. A more expressive cardinal direction calculus can be obtained by using $\mathsf{IA}_7$ — the subalgebra of IA generated by $\mathcal{B}_{int}^{7} = \{\mathsf{b}, (\mathsf{mo}), (\mathsf{sfd}), \mathsf{eq}, (\mathsf{sfd})^\sim, (\mathsf{mo})^\sim, \mathsf{bi}\}$. We define

$$\mathcal{B}_{rec}^{49} = \{\alpha \otimes \beta : \alpha, \beta \in \mathcal{B}_{int}^{7}\}. \quad (13)$$

Clearly, $\mathcal{B}_{rec}^{49}$ is a set of JEPD rectangle relations. We write DIR49 for the Boolean algebra generated by $\mathcal{B}_{rec}^{49}$. As a qualitative calculus, DIR49 is coarser than ERA but finer than DIR9. Later, in Section 7.2, we will show that DIR49 provides a reasonable approximation of ERA.

*Remark* 3.1. One natural requirement for a direction calculus is that it should support definitions of the above four cardinal directions. DIR9 and DIR49 are the two-dimensional counterparts of $\mathcal{B}_3$ and $\mathcal{B}_7$ (see Remark 2.2). These directional calculi do support definitions of the four cardinal directions.

It is worth stressing that these directional calculi — DIR9, DIR49, ERA — are all defined over $U$, the set of bounded regions, where a bounded region may have multiple pieces.

# 4 Combination of Two Qualitative Calculi: The General Case

In this section we consider reasoning problems concerning the combination of two different calculi. The major obstacle is that different kinds of relations may interact with each other. For example, the fact that $a$ is a *part of* $b$ and the fact that $a$ is *larger than* $b$ cannot both be true at the same time.

Suppose $\mathfrak{A}, \mathfrak{B}$ are two qualitative calculi defined on the same universe $D$, and suppose $\mathcal{B}_a$ and $\mathcal{B}_b$ are the sets of basic relations in $\mathfrak{A}$ and $\mathfrak{B}$, respectively. These two calculi describe different kinds of qualitative information of entities in $D$.

Instead of developing a new hybrid calculi, we deal with the reasoning problem directly. Let $\mathcal{N}_a$ and $\mathcal{N}_b$ be two networks of constraints over $\mathfrak{A}$ and $\mathfrak{B}$ which involve the same set of variables. One fundamental reasoning problem for combining $\mathfrak{A}$ and $\mathfrak{B}$ is deciding whether $\mathcal{N}_a \cup \mathcal{N}_b$ is satisfiable. We call this decision problem the *joint satisfaction problem* (JSP) over $\mathfrak{A}$ and $\mathfrak{B}$.

To stress that $\mathcal{N}_a$ and $\mathcal{N}_b$ are defined on the same set of variables, in what follows we write $\mathcal{N}_a \uplus \mathcal{N}_b$, instead of $\mathcal{N}_a \cup \mathcal{N}_b$, for the union of $\mathcal{N}_a$ and $\mathcal{N}_b$.

We next introduce two local constraint propagation techniques in order to provide partial solution to the joint satisfaction problem.

## 4.1 Bi-Closure of Joint Networks

We start with the simplest case where only two variables are involved in $\mathcal{N}_a$ and $\mathcal{N}_b$.

**Definition 4.1.** For a relation $\alpha$ in $\mathfrak{A}$ and a relation $\beta$ in $\mathfrak{B}$, we say $\alpha$ and $\beta$ are *consistent* if the joint network $\{x\alpha y\} \uplus \{x\beta y\}$ has a solution in $D$, i.e. there exist $a, b \in D$ s.t. $a\alpha b$ and $a\beta b$.

*Remark* 4.1. In this paper we do not distinguish between a relation and its model or interpretation in a universe. This is because in most cases we only consider calculi defined on the same universe. Two relations from different calculi interact if they have common instances. The interaction between a basic relation in $\mathfrak{A}$ and a basic relation in $\mathfrak{B}$ is measured in a yes/no fashion. The interaction between a (non-basic) relation in $\mathfrak{A}$ and a (non-basic) relation in $\mathfrak{B}$ will be measured by the notion of bi-closure (see Definition 4.2).

The next lemma follows directly. Note that as relations defined on the same universe, $\alpha$ and $\beta$ may intersect.

**Lemma 4.1.** *For $\alpha$ in $\mathfrak{A}$ and $\beta$ in $\mathfrak{B}$, $\alpha$ and $\beta$ are consistent iff $\alpha \cap \beta \neq \varnothing$.*

Clearly, the universal relation $\top$ is consistent with any nonempty relation $\alpha$ in $\mathfrak{A}$. Moreover, for each nonempty $\alpha$ in $\mathfrak{A}$, there is a smallest relation in $\mathfrak{B}$ which contains $\alpha$. This relation is the largest one in $\mathfrak{B}$ such that $\alpha \cap \beta \neq \varnothing$ but $\alpha \cap -\beta = \varnothing$, where $-\beta$ is the (relation) complement of $\beta$.

**Lemma 4.2.** *Let $\alpha$ be a relation in $\mathfrak{A}$. Then there exists a smallest relation $\beta$ in $\mathfrak{B}$ such that $\alpha$ is consistent with $\beta$ but not consistent with $-\beta$.*

We denote $\mathfrak{B}(\alpha)$ for this relation, and call it the *$\alpha$-induced relation* in $\mathfrak{B}$. Recall that $\mathcal{B}_b$ is the set of basic relations (or atoms) in $\mathfrak{B}$. The $\alpha$-induced relation $\mathfrak{B}(\alpha)$ can be computed as follows.

**Lemma 4.3.** *For a relation $\alpha$ in $\mathfrak{A}$, its induced relation in $\mathfrak{B}$ is the union of all basic relations in $\mathfrak{B}$ that are consistent with $\alpha$, i.e.*

$$\mathfrak{B}(\alpha) = \bigcup\{\beta \in \mathcal{B}_b : \alpha \cap \beta \neq \varnothing\}. \tag{14}$$

Moreover, since $\mathcal{B}_a$ is the set of basic relations (or atoms) in $\mathfrak{A}$, we have

**Lemma 4.4.** *The $\alpha$-induced relation $\mathfrak{B}(\alpha)$ is the union of all $\mathfrak{B}(\alpha')$ with $\alpha' \subseteq \alpha$ and $\alpha' \in \mathcal{B}_a$, i.e.*

$$\mathfrak{B}(\alpha) = \bigcup\{\mathfrak{B}(\alpha') : (\alpha' \in \mathcal{B}_a) \,\&\, (\alpha' \subseteq \alpha)\} \tag{15}$$

$$= \bigcup\{\beta \in \mathcal{B}_b : (\exists \alpha' \in \mathcal{B}_a)[(\alpha' \subseteq \alpha) \,\&\, (\alpha' \cap \beta \neq \varnothing)]\} \tag{16}$$

Given a joint network $\{x\alpha y\} \uplus \{x\beta y\}$, no information will be lost if we subtract from $\beta$ ($\alpha$, resp.) those basic relations that are not consistent with $\alpha$ ($\beta$, resp.). Recall we say two (joint) networks are *equivalent* if they have the same set of solutions.

**Proposition 4.1.** *For a relation $\alpha \in \mathfrak{A}$, and a relation $\beta \in \mathfrak{B}$, $\{x\alpha y\} \uplus \{x\beta y\}$ is equivalent to $\{x\alpha[\beta]y\} \uplus \{x\beta[\alpha]y\}$, i.e. $\alpha[\beta] \cap \beta[\alpha] = \alpha \cap \beta$, where*

$$\alpha[\beta] \equiv \alpha \cap \mathfrak{A}(\beta), \ \beta[\alpha] \equiv \beta \cap \mathfrak{B}(\alpha).$$

*Proof.* To show $\alpha[\beta] \cap \beta[\alpha] = \alpha \cap \beta$, we need only show $\alpha \cap \beta \subseteq \mathfrak{A}(\beta) \cap \mathfrak{B}(\alpha)$. Take $(u,v) \in \alpha \cap \beta$. Suppose $\alpha^*$ and $\beta^*$ are the atomic relations in $\mathfrak{A}$ and, respectively, $\mathfrak{B}$ that contain $(u,v)$. Since $(u,v) \in \beta^* \cap \alpha \neq \varnothing$, by the definition of $\mathfrak{B}(\alpha)$, we know $\beta^* \subseteq \mathfrak{B}(\alpha)$. Hence $(u,v) \in \mathfrak{B}(\alpha)$. Similarly, we know $(u,v) \in \mathfrak{A}(\beta)$. Therefore, $(u,v)$ is an instance of $\mathfrak{A}(\beta) \cap \mathfrak{B}(\alpha)$. Because $(u,v)$ is an arbitrary instance of $\alpha \cap \beta$, we know $\alpha \cap \beta \subseteq \mathfrak{A}(\beta) \cap \mathfrak{B}(\alpha)$ holds. □

In case that $\{x\alpha' y\} \uplus \{x\beta' y\}$ is equivalent to $\{x\alpha y\} \uplus \{x\beta y\}$, we also say $\langle \alpha', \beta' \rangle$ is equivalent to $\langle \alpha, \beta \rangle$. The following lemma shows that $\langle \alpha[\beta], \beta[\alpha] \rangle$ is the smallest pair of constraints which is equivalent to $\langle \alpha, \beta \rangle$.

**Lemma 4.5.** *For $\alpha, \alpha' \in \mathfrak{A}$ and $\beta, \beta' \in \mathfrak{B}$, if $\langle \alpha', \beta' \rangle$ is equivalent to $\langle \alpha, \beta \rangle$, i.e. $\alpha' \cap \beta' = \alpha \cap \beta$, then $\alpha[\beta] \subseteq \alpha'$ and $\beta[\alpha] \subseteq \beta'$.*

*Proof.* Take $(u,v) \in \alpha[\beta] = \alpha \cap \mathfrak{A}(\beta)$. By the definition of $\mathfrak{A}(\beta)$, there exists an $\mathfrak{A}$ atom $\alpha^*$ such that $(u,v) \in \alpha^*$ and $\alpha^* \cap \beta \neq \varnothing$. There must exist a $\mathfrak{B}$ atom $\beta^*$ such that $\beta^* \subseteq \beta$ and $\alpha^* \cap \beta^* \neq \varnothing$. By $(u,v) \in \alpha$, we know $\alpha^*$ is contained in $\alpha$. So we have $\alpha^* \cap \beta^* \subseteq \alpha \cap \beta$. Because $\langle \alpha, \beta \rangle$ is equivalent to $\langle \alpha', \beta' \rangle$, we have $\alpha^* \cap \beta^* \subseteq \alpha' \cap \beta'$. Note that $\alpha^* \cap \alpha' \neq \varnothing$. We know $\alpha^*$, as an $\mathfrak{A}$ atom, is also contained in $\alpha'$. This shows $(u,v)$ is also an instance of $\alpha'$. Therefore, we have $\alpha[\beta] \subseteq \alpha'$. Similarly, we can show $\beta[\alpha] \subseteq \beta'$. □

We say a pair of constraints $\langle \alpha, \beta \rangle$ is *bi-closed* if $\alpha = \alpha[\beta]$ and $\beta = \beta[\alpha]$. It is straightforward to see that $\langle \alpha[\beta], \beta[\alpha] \rangle$ is bi-closed. By Lemma 4.5, it is clear that $\langle \alpha[\beta], \beta[\alpha] \rangle$ is the only bi-closed pair which is equivalent to $\langle \alpha, \beta \rangle$. We call $\langle \alpha[\beta], \beta[\alpha] \rangle$ the *bi-closure* of $\langle \alpha, \beta \rangle$.

The notion of bi-closure can easily be generalized to arbitrary constraint networks.

**Definition 4.2** (bi-closure). For two networks $\mathcal{N}_a = \{v_i \alpha_{ij} v_j\}_{i,j=1}^n$ and $\mathcal{N}_b = \{v_i \beta_{ij} v_j\}_{i,j=1}^n$ over the same $n$ variables, define $\overline{\mathcal{N}}_a = \{v_i \alpha_{ij}[\beta_{ij}] v_j\}_{i,j=1}^n$ and $\overline{\mathcal{N}}_b = \{v_i \beta_{ij}[\alpha_{ij}] v_j\}_{i,j=1}^n$. We call $\overline{\mathcal{N}}_a \uplus \overline{\mathcal{N}}_b$ the *bi-closure* of $\mathcal{N}_a \uplus \mathcal{N}_b$, and say $\mathcal{N}_a \uplus \mathcal{N}_b$ is *bi-closed* if $\overline{\mathcal{N}}_a = \mathcal{N}_a$ and $\overline{\mathcal{N}}_b = \mathcal{N}_b$, i.e. if $\alpha_{ij} = \alpha_{ij}[\beta_{ij}]$ and $\beta_{ij} = \beta_{ij}[\alpha_{ij}]$ for each pair $(i, j)$.

The following lemma shows that $\mathcal{N}_a \uplus \mathcal{N}_b$ and its bi-closure are equivalent, i.e. they have the same set of solutions.

**Lemma 4.6.** *Let $\mathcal{N}_a, \mathcal{N}_b$ and $\overline{\mathcal{N}}_a, \overline{\mathcal{N}}_b$ be as in Definition 4.2. Then $\overline{\mathcal{N}}_a \uplus \overline{\mathcal{N}}_b$ and $\mathcal{N}_a \uplus \mathcal{N}_b$ are equivalent.*

*Proof.* Since $\alpha_{ij}[\beta_{ij}] \subseteq \alpha_{ij}$ and $\beta_{ij}[\alpha_{ij}] \subseteq \beta_{ij}$, we know each solution to the bi-closure is also a solution to $\mathcal{N}_a \uplus \mathcal{N}_b$. On the other hand, suppose $\{a_i\}_{i=1}^n$ is a solution to $\mathcal{N}_a \uplus \mathcal{N}_b$. By Proposition 4.1, $\{v_i \alpha_{ij}[\beta_{ij}] v_j\} \uplus \{v_i \beta_{ij}[\alpha_{ij}] v_j\}$ is equivalent to $\{v_i \alpha_{ij} v_j\} \uplus \{v_i \beta_{ij} v_j\}$. Therefore $(a_i, a_j)$ is also an instance of both $\alpha_{ij}[\beta_{ij}]$ and $\beta_{ij}[\alpha_{ij}]$. This shows that $\{a_i\}_{i=1}^n$ is a solution to $\overline{\mathcal{N}}_a \uplus \overline{\mathcal{N}}_b$. □

It is clear that the bi-closure of a joint network can be computed in $O(n^2)$ time. In what follows, we also call $\overline{\mathcal{N}}_a$ the bi-closure of $\mathcal{N}_a$ w.r.t. $\mathcal{N}_b$, and call $\overline{\mathcal{N}}_b$ the bi-closure of $\mathcal{N}_b$ w.r.t. $\mathcal{N}_a$.

## 4.2 Bipath-Consistency

Gerevini and Renz [12] proposed a cubic local constraint propagation algorithm, termed BIPATH-CONSISTENCY, which is a modification of Allen's path-consistency algorithm (PCA) [1]. BIPATH-CONSISTENCY operates on a graph of constraints, where each edge is labeled by a pair of relations. In our notation, the key updating rules used in BIPATH-CONSISTENCY are

$$\alpha_{ij} \leftarrow \alpha_{ij}[\beta_{ij}] \cap \alpha_{ik}[\beta_{ik}] \circ_w \alpha_{kj}[\beta_{kj}] \tag{17}$$

$$\beta_{ij} \leftarrow \beta_{ij}[\alpha_{ij}] \cap \beta_{ik}[\alpha_{ik}] \circ_w \beta_{kj}[\alpha_{kj}] \tag{18}$$

The next lemma characterizes the output of BIPATH-CONSISTENCY.

**Lemma 4.7.** *For an input joint network $\mathcal{N}_a \uplus \mathcal{N}_b$, suppose BIPATH-CONSISTENCY returns `succeed` and $\mathcal{N}_a' \uplus \mathcal{N}_b'$ is its output. Then $\mathcal{N}_a' \uplus \mathcal{N}_b'$ is bi-closed and $\mathcal{N}_a'$ and $\mathcal{N}_b'$ are path-consistent. On the other hand, if the input $\mathcal{N}_a \uplus \mathcal{N}_b$ is bi-closed and $\mathcal{N}_a$ and $\mathcal{N}_b$ are path-consistent, then BIPATH-CONSISTENCY returns `succeed` and the output joint network is $\mathcal{N}_a \uplus \mathcal{N}_b$ itself.*

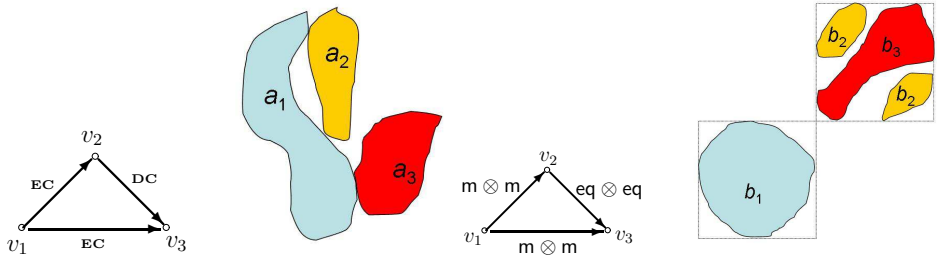

Figure 4: RCC8 network $\mathcal{N}^1_{top}$ and ERA network $\mathcal{N}^1_{dir}$, where $\{a_1, a_2, a_3\}$ is a solution to $\mathcal{N}^1_{top}$, and $\{b_1, b_2, b_3\}$ is a solution to $\mathcal{N}^1_{dir}$, where $b_2$ contains two connected components.

This justifies the rationality of the following definition.

**Definition 4.3.** A joint network $\mathcal{N}_a \uplus \mathcal{N}_b$ is called *bipath-consistent* if it is bi-closed and both $\mathcal{N}_a$ and $\mathcal{N}_b$ are path-consistent.

Clearly, any satisfiable joint network can be transferred to an equivalent bipath-consistent joint network in cubic time using BIPATH-CONSISTENCY. The next subsection shows that there exists a bipath-consistent joint network of basic RCC8 and ERA constraints that is inconsistent.

### 4.3 Bipath-Consistency Is Incomplete for RCC8 and ERA

Suppose $\mathcal{N}_{top} = \{v_i \theta_{ij} v_j\}_{i,j=1}^n$ and $\mathcal{N}_{dir} = \{v_i \delta_{ij} v_j\}_{i,j=1}^n$ are, resp., a topological (RCC8) and a directional (ERA) constraint network over $V = \{v_i\}_{i=1}^n$. Without loss of generality, in the remainder of this paper we assume

(i) $\theta_{ii} = \mathbf{EQ}$ for all $i$, and $\theta_{ij} \neq \mathbf{EQ}$ and $\theta_{ij} = \theta_{ji}^\sim$ for all $i \neq j$; and

(ii) $\delta_{ii} = \mathsf{eq} \otimes \mathsf{eq}$ and $\delta_{ij} = \delta_{ji}^\sim$ for all $i, j$.

The following examples show that a bipath-consistent joint network may be unsatisfiable.

**Example 4.1.** Take $V = \{v_1, v_2, v_3\}$, $\mathcal{N}^1_{top} = \{v_i \theta_{ij} v_j\}_{i,j=1}^3$ and $\mathcal{N}^1_{dir} = \{v_i \delta_{ij} v_j\}_{i,j=1}^3$ are, respectively, the following two networks (see Figure 4):

- $\theta_{12} = \theta_{13} = \mathbf{EC}, \theta_{23} = \mathbf{DC}$;
- $\delta_{12} = \delta_{13} = \mathsf{m} \otimes \mathsf{m}, \delta_{23} = \mathsf{eq} \otimes \mathsf{eq}$.

Since $\{a_1, a_2, a_3\}$ and $\{b_1, b_2, b_3\}$ are, resp., solutions to $\mathcal{N}^1_{top}$ and $\mathcal{N}^1_{dir}$ (see Figure 4), we know these two basic networks are satisfiable and path-consistent. Note that all relations in the two networks are defined over the set of bounded regions. For $\alpha \in \{\mathbf{DC}, \mathbf{EC}\}$ and $\beta \in \{\mathsf{m} \otimes \mathsf{m}, \mathsf{eq} \otimes \mathsf{eq}\}$, it is easy to show that $\alpha \cap \beta$ is nonempty (cf. Lemma 5.2). Therefore, the combined network is bi-closed, hence bipath-consistent by definition. But it is impossible to find a solution to $\mathcal{N}^1_{top} \uplus \mathcal{N}^1_{dir}$. This is because, if $\{a_i^*\}_{i=1}^3$ is a solution of $\mathcal{N}^1_{top} \uplus \mathcal{N}^1_{dir}$, then by $\delta_{23} = \mathsf{eq} \otimes \mathsf{eq}$ and $\delta_{12} = \mathsf{m} \otimes \mathsf{m}$ we know $\mathsf{MBR}(a_2^*) = \mathsf{MBR}(a_3^*)$ and $(\mathsf{MBR}(a_1^*), \mathsf{MBR}(a_2^*)) \in \mathsf{m} \otimes \mathsf{m}$. Write $P$ for the common point of $\mathsf{MBR}(a_1^*)$ and $\mathsf{MBR}(a_2^*)$. Clearly, $a_1^* \cap a_i^* \subseteq \{P\}$ ($i = 2, 3$). By $a_1^* \mathbf{EC} a_i^*$ ($i = 2, 3$) we know $a_1^* \cap a_i^* = \{P\}$. This shows $P \in a_2^* \cap a_3^* \neq \varnothing$, which contradicts with the topological constraint $\theta_{23} = \mathbf{DC}$. Therefore, $\mathcal{N}^1_{top} \uplus \mathcal{N}^1_{dir}$ is bipath-consistent but unsatisfiable.

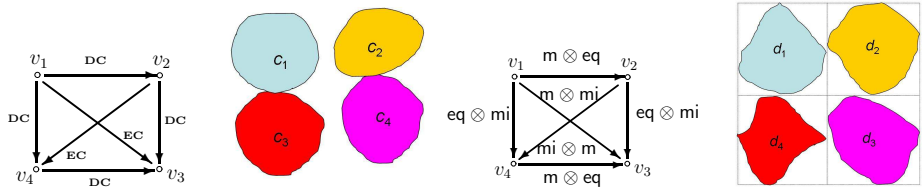

Figure 5: RCC8 network $\mathcal{N}^2_{top}$ and ERA network $\mathcal{N}^2_{dir}$, where $\{c_1, c_2, c_3, c_4\}$ is a solution to $\mathcal{N}^2_{top}$, and $\{d_1, d_2, d_3, d_4\}$ is a solution to $\mathcal{N}^2_{dir}$.

The next example further shows that, even if all sub-networks involving three variables are satisfiable, the joint network may still be unsatisfiable.

**Example 4.2.** Take $V = \{v_i\}_{i=1}^4$, $\mathcal{N}^2_{top}$ and $\mathcal{N}^2_{dir}$ are, respectively, the following networks (see Figure 5).

$$\theta_{ij} = \begin{cases} \mathbf{EC}, & (i,j) = (1,3) \text{ or } (i,j) = (2,4); \\ \mathbf{DC}, & \text{otherwise.} \end{cases}$$

- $\delta_{12} = \mathsf{m} \otimes \mathsf{eq}, \delta_{13} = \mathsf{m} \otimes \mathsf{mi}, \delta_{14} = \mathsf{eq} \otimes \mathsf{mi}$;
- $\delta_{23} = \mathsf{eq} \otimes \mathsf{mi}, \delta_{24} = \mathsf{mi} \otimes \mathsf{mi}, \delta_{34} = \mathsf{mi} \otimes \mathsf{eq}$

It is straightforward to verify that all sub-networks of the joint network $\mathcal{N}^2_{top} \uplus \mathcal{N}^2_{dir}$ which involve three variables are satisfiable.

Since $\{c_1, c_2, c_3, c_4\}$ and $\{d_1, d_2, d_3, d_4\}$ are, resp., solutions to $\mathcal{N}^2_{top}$ and $\mathcal{N}^2_{dir}$ (see Figure 5), the two basic networks are satisfiable and path-consistent. It is also easy to check that $\mathbf{EC}$ and $\mathbf{DC}$ are consistent with all rectangle relations which appear in $\mathcal{N}^2_{dir}$ (cf. Lemma 5.2). Therefore the joint network is bi-closed. But it is impossible to find a solution to $\mathcal{N}^2_{top} \uplus \mathcal{N}^2_{dir}$. This is because by $\theta_{13} = \mathbf{EC}$ and $\delta_{13} = \mathsf{m} \otimes \mathsf{mi}$, we know $v_1$ and $v_3$ must share a unique point $P$. Similarly, $v_2$ and $v_4$ also share a unique point $Q$. It is also clear that $P$ should be identical with $Q$. This suggests that $v_1$ and $v_2$ are externally connected. A contradiction with $\theta_{12} = \mathbf{DC}$.

The above examples show that BIPATH-CONSISTENCY is incomplete for solving the JSP over RCC8 and ERA. In the following sections, we turn to the coarser calculus DIR49. We first show how BIPATH-CONSISTENCY separates topological constraints in some maximal tractable subclasses of RCC8 from directional constraints in DIR49, and then exploit this separation theorem to approximately solve the JSP over RCC8 and ERA. Before this, the next section is devoted to investigating the pairwise interaction between RCC8 and ERA relations.

## 5 Pairwise Interaction between RCC8 and ERA Relations

Given an RCC8 relation $\theta$ and an ERA relation $\delta$, we now consider how to compute $\{v_1\theta[\delta]v_2\} \uplus \{v_1\delta[\theta]v_2\}$, the bi-closure (see Definition 4.2) of $\{v_1\theta v_2\} \uplus \{v_1\delta v_2\}$.

We write $\mathrm{ERA}(\theta)$ for the $\theta$-induced ERA relation and write $\mathrm{RCC}(\delta)$ for the $\delta$-induced RCC8 relation. This means, $\mathrm{ERA}(\theta)$ is the smallest ERA relation which contains $\theta$, and $\mathrm{RCC}(\delta)$ is the smallest RCC8 relation which contains $\delta$ (cf. Lemma 4.2). By Lemma 4.3, we know $\mathrm{ERA}(\theta)$ is the union of all $\mathrm{ERA}(\theta')$, where $\theta'$ is a basic RCC8 relation contained in $\theta$. A similar conclusion holds for $\mathrm{RCC}(\delta)$. Furthermore, by Proposition 4.1, we know $\theta[\delta] = \theta \cap \mathrm{RCC}(\delta)$ and $\delta[\theta] = \delta \cap \mathrm{ERA}(\theta)$. So to compute $\theta[\delta]$ and $\delta[\theta]$ for arbitrary $\theta$ and $\delta$, we first consider the special case when $\theta$ and $\delta$ are basic, and then compute for the general case by using Lemma 4.3 and Proposition 4.1.

Since ERA contains 169 basic rectangle relations, it will be convenient to classify these relations into groups. One natural way is by introducing the following rectangle version of RCC8.

**Definition 5.1** (MRCC8). We say two bounded regions $a, b$ in $U$ are related by **MDC** (**MEC**, **MPO**, **MEQ**, **MTPP**, **MNTPP**, **MTPP**$^\sim$, **MNTPP**$^\sim$, resp.) if **DC** (**EC**, **PO**, **EQ**, **TPP**, **NTPP**, **TPP**$^\sim$, **NTPP**$^\sim$, resp.) is the basic RCC8 relation between $\mathsf{MBR}(a)$ and $\mathsf{MBR}(b)$, the minimum bounding rectangles of $a$ and $b$. We call the qualitative calculus on $U$ generated by

$$\mathcal{B}_{mtop} \equiv \{\mathbf{MDC}, \mathbf{MEC}, \mathbf{MPO}, \mathbf{MEQ}, \mathbf{MTPP}, \mathbf{MNTPP}, \mathbf{MTPP}^\sim, \mathbf{MNTPP}^\sim\} \tag{19}$$

the MRCC8 Algebra.

Proof of the following lemma is straightforward.

**Lemma 5.1.** *Each basic relation in* ERA *is contained in one and only one basic MRCC8 relation. Precisely, for a basic* ERA *relation* $\alpha \otimes \beta$, *we have*

1. *if* $\alpha \otimes \beta = \mathsf{eq} \otimes \mathsf{eq}$, *then* $\alpha \otimes \beta = \mathbf{MEQ}$;

2. *if* $\alpha \otimes \beta = \mathsf{d} \otimes \mathsf{d}$, *then* $\alpha \otimes \beta = \mathbf{MNTPP}$;

3. *if* $\alpha \otimes \beta = \mathsf{di} \otimes \mathsf{di}$, *then* $\alpha \otimes \beta = \mathbf{MNTPP}^\sim$;

4. *else if* $\alpha, \beta \in \{\mathsf{s}, \mathsf{d}, \mathsf{f}, \mathsf{eq}\}$, *then* $\alpha \otimes \beta \subset \mathbf{MTPP}$;

Figure 6: Amalgamation of basic rectangle relations, where Q, T, Ti, N, Ni represent **MEQ**, **MTPP**$^\sim$, **MTPP**$^\sim$, **MNTPP**, and **MNTPP**$^\sim$, respectively.

5. *else if* $\alpha, \beta \in \{$si, di, fi, eq$\}$, *then* $\alpha \otimes \beta \subset$ **MTPP**$^\sim$;

6. *else if* $\alpha \in \{$b, bi$\}$ *or* $\beta \in \{$b, bi$\}$, *then* $\alpha \otimes \beta \subset$ **MDC**;

7. *else if* $\alpha \in \{$m, mi$\}$ *or* $\beta \in \{$m, mi$\}$, *then* $\alpha \otimes \beta \subset$ **MEC**;

8. *else* $\alpha \otimes \beta \subset$ **MPO**.

Take the first and the last items as examples. For two bounded regions $a, b$, item 1 is equivalent to saying that $(\mathsf{MBR}(a), \mathsf{MBR}(b))$ is an instance of eq $\otimes$ eq iff it is an instance of **MEQ**, i.e. $\mathsf{MBR}(a) = \mathsf{MBR}(b)$. Item 8 states that if the basic ERA relation between $\mathsf{MBR}(a)$ and $\mathsf{MBR}(b)$ does not satisfy the precondition of items 1-7, then $\mathsf{MBR}(a)$ must partially overlap $\mathsf{MBR}(b)$. In what follows, we call a basic ERA relation an **MDC** relation, if it is contained in **MDC**, and similarly for relations contained in **MEC**, **MPO**, etc.

The next lemma summarizes the $\theta$-induced ERA relations, $\mathrm{ERA}(\theta)$, for all basic RCC8 relations $\theta$. Recall that $\mathrm{ERA}(\theta)$ is, by definition, the smallest ERA relation which contains $\theta$.

**Lemma 5.2.** *For a basic RCC8 relation $\theta$, the $\theta$-induced* ERA *relation* $\mathrm{ERA}(\theta)$ *is as follows:*

1. $\mathrm{ERA}(\mathbf{EQ}) = $ eq $\otimes$ eq;

2. $\mathrm{ERA}(\mathbf{NTPP}) = $ d $\otimes$ d;

3. $\mathrm{ERA}(\mathbf{NTPP}^\sim) = $ di $\otimes$ di;

4. $\mathrm{ERA}(\mathbf{TPP}) = $ (sdfeq) $\otimes$ (sdfeq);

5. $\mathrm{ERA}(\mathbf{TPP}^\sim) = $ (sdfeq)$^\sim$ $\otimes$ (sdfeq)$^\sim$;

6. $\mathrm{ERA}(\mathbf{DC})$ *is the union of all* ERA *basic relations, i.e.* $\mathrm{ERA}(\mathbf{DC}) = \top$;

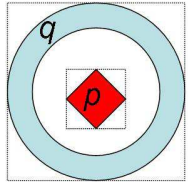

Figure 7: Illustrations of two connected regions $p, q$ and their minimum bounding rectangles.

7. ERA(**EC**) *is the union of all* ERA *basic relations that are not* **MDC** *relations;*

8. ERA(**PO**) *is the union of all* ERA *relations that are neither* **MDC** *nor* **MEC** *relations.*

*Proof.* We take the case when $\theta = $ **TPP** as an example; the others are similar. Suppose $a, b$ are two bounded regions such that $a$**TPP**$b$. We show (MBR($a$), MBR($b$)) $\in$ (sdfeq) $\otimes$ (sdfeq). Write $I_x(a)$ and $I_x(b)$ for the $x$-projections (cf. Figure 1) of $a$ and $b$, resp. By $a$**TPP**$b$, we know $a \subset b$. It is clear that $I_x(a) \subseteq I_x(b)$. This is equivalent to saying that the interval relation between $I_x(a)$ and $I_x(b)$ is (sdfeq). The same IA relation also holds for the $y$-projections of $a$ and $b$. Recall that MBR($a$) = $I_x(a) \times I_y(a)$ and MBR($b$) = $I_x(b) \times I_y(b)$. We have (MBR($a$), MBR($b$)) $\in$ (sdfeq) $\otimes$ (sdfeq). By the definition of the extended rectangle relations, $(a, b)$ is an instance of the ERA relation (sdfeq) $\otimes$ (sdfeq). Therefore **TPP** is contained in (sdfeq) $\otimes$ (sdfeq). We next show this is also the smallest ERA relation which contains **TPP**. To this end, we need to show **TPP** is consistent with each rectangle relation $\alpha \otimes \beta$ with $\alpha, \beta \in \{s, d, f, eq\}$. Take $d \otimes d$ and $eq \otimes eq$ as examples. Figure 7 shows two connected regions $p$ and $q$. Let $r = p \cup q$. Then MBR($r$) = MBR($q$), and (MBR($p$), MBR($r$)) $\in d \otimes d$. In other words, $(p, r)$ is an instance of the ERA relation $d \otimes d$, and $(q, r)$ is an instance of the ERA relation $eq \otimes eq$. It is also clear that $p$ and $q$ are two tangential proper parts of $r$, i.e. $p$**TPP**$r$, $q$**TPP**$r$. □

As a corollary, we have

**Corollary 5.1.** *For any RCC8 relation $\theta$, we have*

- *If $\theta \cap$ **DC** $= \varnothing$, then* ERA($\theta$) *contains no* **MDC** *relation.*

- *If* **TPP** $\subseteq \theta \subseteq$ **P**, *then* ERA($\theta$) = (sdfeq) $\otimes$ (sdfeq),

*where* **P** *is the union of* **TPP**, **NTPP**, *and* **EQ**.

*Proof.* This is because ERA($\theta$) is the union of all ERA($\theta'$), where $\theta'$ is a basic RCC8 relation that is contained in $\theta$. The conclusions then follow directly from Lemma 5.2. □

Just like Lemma 5.2, the next lemma summarizes the $\delta$-induced RCC8 relations, RCC($\delta$), for all basic ERA relations $\delta$. Recall that RCC($\delta$) is the smallest RCC8 relation which contains $\delta$.

**Lemma 5.3.** *For a basic ERA relation $\delta$, the $\delta$-induced RCC8 relation* RCC($\delta$) *is as follows:*

1. RCC($\delta$) = **DC** *if $\delta$ is an* **MDC** *relation;*

2. RCC($\delta$) = **DC** ∪ **EC** *if $\delta$ is an* **MEC** *relation;*

3. RCC($\delta$) = **DC** ∪ **EC** ∪ **PO** *if $\delta$ is an* **MPO** *relation;*

4. RCC($\delta$) = **DC** ∪ **EC** ∪ **PO** ∪ **TPP** *if $\delta$ is an* **MTPP** *relation;*

5. RCC($\delta$) = **DC**∪**EC**∪**PO**∪**TPP**∪**NTPP** *if $\delta$ is an* **MNTPP** *relation;*

6. RCC($\delta$) = **DC** ∪ **EC** ∪ **PO** ∪ **TPP**$^\sim$ *if $\delta$ is an* **MTPP**$^\sim$ *relation;*

7. RCC($\delta$) = **DC** ∪ **EC** ∪ **PO** ∪ **TPP**$^\sim$ ∪ **NTPP**$^\sim$ *if $\delta$ is an* **MNTPP**$^\sim$ *relation;*

8. RCC($\delta$) = **DC**∪**EC**∪**PO**∪**EQ**∪**TPP**∪**TPP**$^\sim$ *if $\delta$ is the* **MEQ** *relation.*

The proof of this lemma is straightforward. We only give some explanation here. The first item states that if $a$**MDC**$b$, i.e. MBR($a$)**DC**MBR($b$), then we should also have $a$**DC**$b$; the last item states that if $a$**MEQ**$b$, i.e. MBR($a$) = MBR($b$), then $a$ and $b$ could be related by any basic RCC8 relation other than **NTPP** and its converse.

# 6 Combining Topological and Directional Constraints

We continue our discussion of the combination of RCC8 and ERA. Recall that we have shown in Section 5.2 that BIPATH-CONSISTENCY is incomplete for determining the joint satisfaction problem (JSP) over RCC8 and ERA. In this section, we adopt DIR49 as our constraint language for directional information, and show $\widehat{\mathcal{H}}_8$ is separable from DIR49, where $\widehat{\mathcal{H}}_8$ is the maximal tractable subclass of RCC8 found in [35]. In this case, we even do not need to call the full BIPATH-CONSISTENCY algorithm.

Given $\mathcal{N}_{top} = \{v_i \theta_{ij} v_j\}_{i,j=1}^n$ and $\mathcal{N}_{dir} = \{v_i \delta_{ij} v_j\}_{i,j=1}^n$, we first compute the bi-closure of $\mathcal{N}_{top} \uplus \mathcal{N}_{dir}$. For convenience, we set $\overline{\theta}_{ij} = \theta_{ij}[\delta_{ij}]$ and $\overline{\delta}_{ij} = \delta_{ij}[\theta_{ij}]$, and let $\overline{\mathcal{N}}_{top} = \{v_i \overline{\theta}_{ij} v_j\}_{i,j=1}^n$ and $\overline{\mathcal{N}}_{dir} = \{v_i \overline{\delta}_{ij} v_j\}_{i,j=1}^n$. We stress that $\overline{\delta}_{ij}$ may be an ERA relation outside DIR49. For example, set $\delta_{ij} = (\mathsf{sfd}) \otimes (\mathsf{sfd})$ and $\theta_{ij} = $ **NTPP**. Then $\overline{\delta}_{ij} = \mathsf{d} \otimes \mathsf{d}$ is outside DIR49. On the other hand, if $\mathcal{N}_{top}$ is over $\widehat{\mathcal{H}}_8$, then each constraint in $\overline{\mathcal{N}}_{top}$ is in $\widehat{\mathcal{H}}_8$. This is because (see Lemma 5.3) RCC($\delta$) is in $\widehat{\mathcal{H}}_8$ for any ERA relation $\delta$, and that $\widehat{\mathcal{H}}_8$ is closed under intersection.

By Lemma 4.6, we know $\mathcal{N}_{top} \uplus \mathcal{N}_{dir}$ and its bi-closure are equivalent.

**Lemma 6.1.** *For an RCC8 network $\mathcal{N}_{top}$ and an ERA network $\mathcal{N}_{dir}$, the joint network $\mathcal{N}_{top} \uplus \mathcal{N}_{dir}$ is satisfiable if and only if its bi-closure $\overline{\mathcal{N}}_{top} \uplus \overline{\mathcal{N}}_{dir}$ is satisfiable.*

In the remainder of this section, we show that if $\mathcal{N}_{top}$ is a path-consistent RCC8 network over $\widehat{\mathcal{H}}_8$ and $\mathcal{N}_{dir}$ is a DIR49 network, then $\mathcal{N}_{top} \uplus \mathcal{N}_{dir}$ is satisfiable if and only if $\overline{\mathcal{N}}_{top}$ and $\overline{\mathcal{N}}_{dir}$ are, independently, satisfiable. To this end, we choose an appropriate scenario $\mathcal{N}^*_{top}$ of $\overline{\mathcal{N}}_{top}$ and an appropriate scenario $\mathcal{N}^*_{dir}$ of $\overline{\mathcal{N}}_{dir}$, and show that $\mathcal{N}^*_{top} \uplus \mathcal{N}^*_{dir}$ is satisfiable. Recall a scenario of $\overline{\mathcal{N}}_{top}$ ($\overline{\mathcal{N}}_{dir}$, resp.) is a basic RCC8 (ERA, resp.) network that refines $\overline{\mathcal{N}}_{top}$ ($\overline{\mathcal{N}}_{dir}$, resp.)

Before constructing $\mathcal{N}^*_{top}$ and $\mathcal{N}^*_{dir}$, we set a condition that they should satisfy.

## 6.1 Compatible Rectangles

Given an RCC8 basic network $\mathcal{N}_{top} = \{v_i \theta_{ij} v_j\}_{i,j=1}^n$, we know $\mathcal{N}_{top}$ is satisfiable if it is path-consistent. Moreover, a solution by bounded regions can be constructed in cubic time [32, 20]. Suppose $\{r_i\}_{i=1}^n$ is a collection of rectangles. We are interested in knowing if there is a solution $\{a_i\}_{i=1}^n$ for $\mathcal{N}_{top}$ such that each $a_i$ is exactly bounded by the rectangle $r_i$. We find a sufficient condition for this question.

**Definition 6.1.** A collection of rectangles $\{r_i\}_{i=1}^n$ are *compatible* with an RCC8 basic network $\mathcal{N}_{top} = \{v_i \theta_{ij} v_j\}_{i,j=1}^n$ if for any $i, j$ we have

- If $\theta_{ij} \neq \mathbf{DC}$, then $r_i \cap r_j$ is a rectangle, i.e. the interior of $r_i \cap r_j$ is nonempty;

- If $\theta_{ij} = \mathbf{TPP}$, then $(r_i, r_j)$ is in $\mathsf{d} \otimes \mathsf{eq}$ or $\mathsf{d} \otimes \mathsf{d}$ or $\mathsf{eq} \otimes \mathsf{d}$ or $\mathsf{eq} \otimes \mathsf{eq}$;

- If $\theta_{ij} = \mathbf{NTPP}$, then $r_i$ is contained in the interior of $r_j$, i.e. $(r_i, r_j) \in \mathsf{d} \otimes \mathsf{d}$;

- If $\theta_{ij} = \mathbf{EQ}$, then $r_i = r_j$.

At first glance, the notion of compatible rectangles seems very strong. For two rectangles $r_i$ and $r_j$, it requires the $x$- or $y$-projections of $r_i$ and $r_j$ not to be related by the IA relations *meet, start, finish,* nor by their converses. The following theorem partially justifies the appropriateness of the notion, where $\{v_i \alpha_{ij} v_j\}_{i=1}^n$ is a scenario of a network $\{v_i \beta_{ij} v_j\}_{i=1}^n$ in a qualitative calculus $\mathfrak{A}$ if $\alpha_{ij}$ is a basic relation in $\mathfrak{A}$ which is contained in $\beta_{ij}$.

**Theorem 6.1.** *Let $\mathcal{N}_{top}$ be an RCC8 network, and let $\mathcal{N}_{dir}$ be a DIR49 network. Suppose $\overline{\mathcal{N}}_{dir}$ is satisfiable. Then $\overline{\mathcal{N}}_{dir}$ has a satisfiable scenario $\mathcal{N}'_{dir} = \{v_i \delta'_{ij} v_j\}_{i,j=1}^n$ such that each $\delta'_{ij}$ has the form $\beta^x_{ij} \otimes \beta^y_{ij}$, where $\beta^x_{ij}, \beta^y_{ij} \in \{\mathsf{b}, \mathsf{o}, \mathsf{d}, \mathsf{eq}, \mathsf{di}, \mathsf{oi}, \mathsf{bi}\}$.*

*Proof.* See Appendix A. □

The next theorem confirms that, for a satisfiable basic RCC8 network $\mathcal{N}_{top}$, we can first find an approximate solution by using rectangles $\{r_i\}_{i=1}^n$, and then get the exact solution $\{a_i^*\}_{i=1}^n$ such that each $a_i^*$ is exactly bounded by $r_i$, i.e. $\mathsf{MBR}(a_i^*) = r_i$.

**Theorem 6.2.** *Let $\mathcal{N}_{top} = \{v_i \theta_{ij} v_j\}_{i,j=1}^n$ be a satisfiable basic RCC8 network. Suppose $\{r_i\}_{i=1}^n$ is a collection of rectangles that are compatible with $\mathcal{N}_{top}$. Then we have a solution $\{a_i^*\}_{i=1}^n$ of $\mathcal{N}_{top}$ such that each $a_i^*$ is a bounded region and $\mathsf{MBR}(a_i^*) = r_i$ for any $1 \leq i \leq n$.*

*Proof.* The proof is similar to that given for RCC8 in [20]. We defer it to Appendix B. □

## 6.2 Separating $\widehat{\mathcal{H}}_8$ from DIR49

In this subsection we prove the separation theorem for $\widehat{\mathcal{H}}_8$ and DIR49. Let $\mathcal{N}_{top} = \{v_i \theta_{ij} v_j\}_{i,j=1}^n$ be a path-consistent RCC8 network over $\widehat{\mathcal{H}}_8$, and let $\mathcal{N}_{dir} = \{v_i \delta_{ij} v_j\}_{i,j=1}^n$ be a DIR49 network. Suppose $\overline{\mathcal{N}}_{top}$ and $\overline{\mathcal{N}}_{dir}$ are satisfiable. We construct an RCC8 basic network $\mathcal{N}_{top}^*$ that refines $\overline{\mathcal{N}}_{top}$. Then we show there is a basic ERA network $\mathcal{N}_{dir}^*$ such that

- $\mathcal{N}_{dir}^*$ refines $\overline{\mathcal{N}}_{dir}$; and
- $\mathcal{N}_{dir}^*$ has a rectangle solution $\{r_i\}_{i=1}^n$ which is compatible with $\mathcal{N}_{top}^*$.

By Theorem 6.2 we know $\mathcal{N}_{top}^* \uplus \mathcal{N}_{dir}^*$, hence $\mathcal{N}_{top} \uplus \mathcal{N}_{dir}$, is satisfiable.

We use the quadratic algorithm proposed by Renz [33] to construct $\mathcal{N}_{top}^*$. For each relation $\theta$ in $\widehat{\mathcal{H}}_8$, we assign a basic relation $\hbar(\theta)$ as follows:

$$\hbar : \widehat{\mathcal{H}}_8 \to \mathcal{B}_{top} \qquad (20)$$

$$\hbar(\theta) = \begin{cases} \mathbf{DC}, & \text{if } \mathbf{DC} \subseteq \theta; \\ \mathbf{EC}, & \text{else if } \mathbf{EC} \subseteq \theta; \\ \mathbf{PO}, & \text{else if } \mathbf{PO} \subseteq \theta; \\ \mathbf{TPP}, & \text{else if } \mathbf{TPP} \subseteq \theta; \\ \mathbf{TPP}^\sim, & \text{else if } \mathbf{TPP}^\sim \subseteq \theta; \\ \theta, & \text{else.} \end{cases}$$

**Lemma 6.2** ([33]). *Let $\mathcal{N}_{top}$ be a path-consistent network over $\widehat{\mathcal{H}}_8$. Then the basic RCC8 network $\mathcal{N}_{top}^* = \{v_i \hbar(\theta_{ij}) v_j\}_{i,j=1}^n$ is satisfiable.*

We next show that the satisfiable RCC8 basic network $\mathcal{N}_{top}^*$ also refines $\overline{\mathcal{N}}_{top}$. To this end, we need the following lemma.

**Lemma 6.3.** *For an RCC8 relation $\theta \in \widehat{\mathcal{H}}_8$ and a DIR49 relation $\delta$, if $\theta[\delta] \neq \varnothing$ and $\delta[\theta] \neq \varnothing$, then $\hbar(\theta) = \hbar(\theta[\delta])$.*

*Proof.* See Appendix C. □

As a corollary, we have

**Lemma 6.4.** *Let $\mathcal{N}_{top} = \{v_i\theta_{ij}v_j\}_{i,j=1}^n$ be a path-consistent RCC8 network over $\widehat{\mathcal{H}}_8$, and let $\mathcal{N}_{dir} = \{v_i\delta_{ij}v_j\}_{i,j=1}^n$ be a DIR49 network. Write $\mathcal{N}_{top}^*$ for the scenario of $\mathcal{N}_{top}$ as constructed in Lemma 6.2. Suppose $\overline{\mathcal{N}}_{top}$ and $\overline{\mathcal{N}}_{dir}$ are satisfiable. Then $\mathcal{N}_{top}^*$ is also a scenario of $\overline{\mathcal{N}}_{top}$.*

By the above lemma, it is easy to see that $\overline{\mathcal{N}}_{top}$ is satisfiable if and only if $\mathcal{N}_{top}^*$ is one of its scenarios. Having found a satisfiable scenario for $\overline{\mathcal{N}}_{top}$, we next show that there is a rectangle solution to $\overline{\mathcal{N}}_{dir}$ that is compatible with $\overline{\mathcal{N}}_{top}$.

**Lemma 6.5.** *For $\mathcal{N}_{top}$, $\mathcal{N}_{dir}$, and $\mathcal{N}_{top}^*$ as above. If $\overline{\mathcal{N}}_{dir}$ is satisfiable, then it has a rectangle solution $\{r_i\}_{i=1}^n$ that is compatible with $\mathcal{N}_{top}^*$.*

*Proof.* By Theorem 6.1 we know $\overline{\mathcal{N}}_{dir}$ has a satisfiable scenario $\mathcal{N}_{dir}^* = \{v_i\delta_{ij}^*v_j\}$ such that each $\delta_{ij}^*$ has the form $\alpha \otimes \beta$ with $\alpha, \beta \in \{\mathsf{b}, \mathsf{o}, \mathsf{d}, \mathsf{eq}, \mathsf{di}, \mathsf{oi}, \mathsf{bi}\}$.

Suppose $\mathcal{I} = \{r_i\}_{i=1}^n$ is a rectangle solution of $\mathcal{N}_{dir}^*$. Clearly, no two rectangles in $\mathcal{I}$ meet at boundaries, i.e. $(r_i, r_j) \notin \mathbf{EC}$ for all $i, j$. In other words, for $r_i$ and $r_j$ in $\mathcal{I}$, we have either $r_i \cap r_j = \varnothing$ or $r_i \cap r_j$ is a rectangle.

We show $\mathcal{I}$ is compatible with $\mathcal{N}_{top}^*$. To this end, we need to show that $\mathcal{I}$ satisfies the four conditions listed in Definition 6.1. Note that $(r_i, r_j)$ is an instance of $\delta_{ij}^* \subseteq \overline{\delta}_{ij} \subseteq \mathrm{ERA}(\theta_{ij})$.

- If $\theta_{ij}^* \neq \mathbf{DC}$, then $\theta_{ij} \cap \mathbf{DC} = \varnothing$. By Corollary 5.1, no basic rectangle relation contained in $\mathrm{ERA}(\theta_{ij})$ is an **MDC** relation. Therefore, by $(r_i, r_j) \in \delta_{ij}^* \subseteq \mathrm{ERA}(\theta_{ij})$ we know $r_i \cap r_j$ is nonempty, hence a rectangle.

- If $\theta_{ij}^* = \mathbf{TPP}$, then $\mathbf{TPP} \subseteq \theta_{ij} \subseteq \mathbf{P}$. By Corollary 5.1, $\mathrm{ERA}(\theta_{ij}) = (\mathsf{sdfeq}) \otimes (\mathsf{sdfeq})$. By the property of $\delta_{ij}^*$ and $(r_i, r_j) \in \delta_{ij}^* \subseteq \mathrm{ERA}(\theta_{ij})$, we know $(r_i, r_j)$ must be an instance of one of the four rectangle relations $\mathsf{d} \otimes \mathsf{eq}$, $\mathsf{d} \otimes \mathsf{d}$, $\mathsf{eq} \otimes \mathsf{d}$, or $\mathsf{eq} \otimes \mathsf{eq}$.

- If $\theta_{ij}^* = \mathbf{NTPP}$, then $\theta_{ij} = \mathbf{NTPP}$. By Lemma 5.2, $\mathrm{ERA}(\mathbf{NTPP}) = \mathsf{d} \otimes \mathsf{d}$. Since $(r_i, r_j) \in \delta_{ij}^*$, we also have $(r_i, r_j) \in \mathsf{d} \otimes \mathsf{d}$.

- If $\theta_{ij}^* = \mathbf{EQ}$, then $\theta_{ij} = \mathbf{EQ}$. By Lemma 5.2, $\mathrm{ERA}(\mathbf{EQ}) = \mathsf{eq} \otimes \mathsf{eq}$. Since $(r_i, r_j) \in \delta_{ij}^*$, we also have $(r_i, r_j) \in \mathsf{eq} \otimes \mathsf{eq}$, i.e. $r_i = r_j$.

Therefore, $\mathcal{I}$ is a rectangle solution of $\overline{\mathcal{N}}_{dir}$ that is compatible with $\mathcal{N}_{top}^*$. □

As a consequence of the above results, we have the following theorem.

**Theorem 6.3.** *Let $\mathcal{N}_{top} = \{v_i\theta_{ij}v_j\}_{i,j=1}^n$ be a path-consistent RCC8 network over $\widehat{\mathcal{H}}_8$, and let $\mathcal{N}_{dir} = \{v_i\delta_{ij}v_j\}_{i,j=1}^n$ be a DIR49 network. Then $\mathcal{N}_{top} \uplus \mathcal{N}_{dir}$ is satisfiable iff $\overline{\mathcal{N}}_{top}$ and $\overline{\mathcal{N}}_{dir}$ are independently satisfiable.*

*Proof.* Suppose $\overline{\mathcal{N}}_{top}$ and $\overline{\mathcal{N}}_{dir}$ are satisfiable. Since $\mathcal{N}_{top}$ is a path-consistent network over $\widehat{\mathcal{H}}_8$, we can construct a basic RCC8 network $\mathcal{N}^*_{top} = \{v_i \theta^*_{ij} v_j\}^n_{i,j=1}$ as in Lemma 6.2. By Lemma 6.4 we know $\mathcal{N}^*_{top}$ is a scenario of $\overline{\mathcal{N}}_{top}$, i.e. $\theta^*_{ij}$ is contained in $\theta_{ij}[\delta_{ij}]$ for all $i, j$.

By Lemma 6.5 we know $\overline{\mathcal{N}}_{dir}$ has a solution $\{r_i\}^n_{i=1}$ that is compatible with $\mathcal{N}^*_{top}$. In other words, $\mathcal{N}^*_{dir}$ and $\{r_i\}^n_{i=1}$ satisfy the conditions of Definition 6.1. Therefore, by Theorem 6.2, we can find a solution $\{c_i\}^n_{i=1}$ of $\mathcal{N}^*_{top}$ which satisfies $\mathsf{MBR}(c_i) = r_i$ for $i = 1, \cdots, n$. So $\{c_i\}^n_{i=1}$ is also a solution of $\overline{\mathcal{N}}_{dir}$. Therefore, $\mathcal{N}_{top} \uplus \mathcal{N}_{dir}$ is satisfiable. □

*Remark* 6.1. For a path-consistent RCC8 network $\mathcal{N}_{top}$ over $\widehat{\mathcal{H}}_8$ and a DIR49 network $\mathcal{N}_{dir}$, to determine if the joint network $\mathcal{N}_{top} \uplus \mathcal{N}_{dir}$ is satisfiable, by the above theorem, we first compute $\overline{\mathcal{N}}_{top}$ and $\overline{\mathcal{N}}_{dir}$, and then check if they are satisfiable independently. Ideally, we wish $\overline{\mathcal{N}}_{dir}$ is also a DIR49 network. But by applying the rules like "**NTPP** enforces $\mathsf{d} \otimes \mathsf{d}$" (Lemma 5.2) constraints in $\overline{\mathcal{N}}_{dir}$ may be outside DIR49. This is not a problem. What we want is to solve the joint constraint network efficiently and do not care how and in which calculus the problem is solved.

By using the rules like "**NTPP** enforces $\mathsf{d} \otimes \mathsf{d}$," we obtain the bi-closure of a joint network. Then, we need only compute if the two separated networks are satisfied independently. This reasoning process is carried in RCC8 and in ERA. Note that there are complete methods for solving the satisfaction problem in both RCC8 and ERA. The joint satisfaction problem defined over $\widehat{\mathcal{H}}_8$ and DIR49 could therefore be solved by Theorem 6.3.

For an RCC8 network $\mathcal{N}_{top}$ over $\widehat{\mathcal{H}}_8$ and a DIR49 network $\mathcal{N}_{dir}$, recall that $\mathcal{N}_{top} \uplus \mathcal{N}_{dir}$ is bipath-consistent if and only if it is bi-closed and both $\mathcal{N}_{top}$ and $\mathcal{N}_{dir}$ are path-consistent. Moreover, if $\mathcal{N}_{top} \uplus \mathcal{N}_{dir}$ is bi-closed, then $\overline{\mathcal{N}}_{top} = \mathcal{N}_{top}$ and $\overline{\mathcal{N}}_{dir} = \mathcal{N}_{dir}$.

The following theorem shows that BIPATH-CONSISTENCY separates $\widehat{\mathcal{H}}_8$ and DIR49.

**Theorem 6.4.** *For an RCC8 network $\mathcal{N}_{top}$ over $\widehat{\mathcal{H}}_8$ and a DIR49 network $\mathcal{N}_{dir}$, suppose $\mathcal{N}'_{top} \uplus \mathcal{N}'_{dir}$ is a bipath-consistent joint network that is equivalent to $\mathcal{N}_{top} \uplus \mathcal{N}_{dir}$. Then $\mathcal{N}_{top} \uplus \mathcal{N}_{dir}$ is satisfiable if $\mathcal{N}'_{top}$ and $\mathcal{N}'_{dir}$ are independently satisfiable.*

*Proof.* Since constraints in $\mathcal{N}'_{dir}$ may be outside DIR49, we cannot apply Theorem 6.3 directly. But $\mathcal{N}'_{top}$ and $\mathcal{N}_{dir}$ satisfy the condition of Theorem 6.3. This means $\mathcal{N}'_{top} \uplus \mathcal{N}_{dir}$ is satisfiable if and only of the two component networks of its bi-closure are independently satisfiable.

We next compute the bi-closure of $\mathcal{N}'_{top} \uplus \mathcal{N}_{dir}$. Suppose $\mathcal{N}_{top} = \{\theta_{ij}\}^n_{i,j=1}$, $\mathcal{N}_{dir} = \{\delta_{ij}\}^n_{i,j=1}$, and $\mathcal{N}'_{top} = \{\theta'_{ij}\}^n_{i,j=1}$, $\mathcal{N}'_{dir} = \{\delta'_{ij}\}^n_{i,j=1}$. We have $\mathcal{N}'_{top} \uplus \mathcal{N}'_{dir}$ is bi-closed due to its bipath-consistency. This means that $\theta'_{ij} = \theta'_{ij}[\delta'_{ij}]$ and $\delta'_{ij} = \delta'_{ij}[\theta'_{ij}]$ for any $i, j$. Note that $\theta'_{ij} \subseteq \theta_{ij}$ and $\delta'_{ij} \subseteq \delta_{ij}$ for any $i, j$. We have

$$\theta'_{ij} = \theta'_{ij}[\delta'_{ij}] = \theta'_{ij} \cap \mathrm{RCC}(\delta'_{ij}) \subseteq \theta'_{ij} \cap \mathrm{RCC}(\delta_{ij}) = \theta'_{ij}[\delta_{ij}] \quad (21)$$

$$\delta'_{ij} = \delta'_{ij}[\theta'_{ij}] = \delta'_{ij} \cap \mathrm{ERA}(\theta'_{ij}) \subseteq \delta_{ij} \cap \mathrm{ERA}(\theta'_{ij}) = \delta_{ij}[\theta'_{ij}]. \quad (22)$$

Set $\widetilde{\mathcal{N}}_{top} = \{\theta'_{ij}[\delta_{ij}]\}_{i,j=1}^n$ and $\widetilde{\mathcal{N}}_{dir} = \{\delta_{ij}[\theta'_{ij}]\}_{i,j=1}^n$. Clearly, $\widetilde{\mathcal{N}}_{top} \uplus \widetilde{\mathcal{N}}_{dir}$ is the bi-closure of $\mathcal{N}'_{top} \uplus \mathcal{N}'_{dir}$. By Equations 21 and 22 we know $\mathcal{N}'_{top}$ refines $\widetilde{\mathcal{N}}_{top}$ and $\mathcal{N}'_{dir}$ refines $\widetilde{\mathcal{N}}_{dir}$. Under the assumption that $\mathcal{N}'_{top}$ and $\mathcal{N}'_{dir}$ are satisfiable, we know $\widetilde{\mathcal{N}}_{top}$ and $\widetilde{\mathcal{N}}_{dir}$ are satisfiable. By Theorem 6.3, this implies $\mathcal{N}'_{top} \uplus \mathcal{N}'_{dir}$, hence $\mathcal{N}_{top} \uplus \mathcal{N}_{dir}$, is satisfiable. □

Recall that applying PCA is sufficient for deciding satisfiability for the RCC8 subclass $\widehat{\mathcal{H}}_8$, and for the ERA subclass $\mathcal{H} \otimes \mathcal{H}$, where $\mathcal{H}$ is the ORD-Horn subclass of IA. We have the following corollary.

**Corollary 6.1.** *Let $\mathcal{N}_{top}$ be an RCC8 network over $\widehat{\mathcal{H}}_8$, and let $\mathcal{N}_{dir}$ be a DIR49 network over $\mathcal{H}_7 \otimes \mathcal{H}_7$, where $\mathcal{H}_7$ is the intersection of $\mathcal{H}$ and the interval algebra $\mathsf{IA}_7$. Then deciding the satisfiability of $\mathcal{N}_{top} \uplus \mathcal{N}_{dir}$ is of cubic complexity.*

*Proof.* It is of quadratic complexity to compute $\mathcal{N}^*_{top}$ and $\overline{\mathcal{N}}_{dir}$. Note that $\overline{\mathcal{N}}_{dir}$ is a rectangle network over $\mathcal{H} \otimes \mathcal{H}$, and applying PCA in RCC8 and ERA is of cubic complexity. □

# 7 Further Discussions

In this section we show how the above separation theorem can be exploited to solve the general joint satisfaction problem over RCC8 and ERA.

## 7.1 Beyond $\widehat{\mathcal{H}}_8$

Theorem 6.3 requires that all topological constraints are in $\widehat{\mathcal{H}}_8$, which is one of the three maximal tractable subclasses ($\widehat{\mathcal{H}}_8$, $\mathcal{Q}_8$, $\mathcal{C}_8$) identified in [33].

For $\mathcal{Q}_8$, a separation theorem can be obtained in a similar way. Given a path-consistent RCC8 network $\mathcal{N}_{top}$ over $\mathcal{Q}_8$, and a DIR49 network $\mathcal{N}_{dir}$, let $\mathcal{N}^*_{top}$ be the scenario of $\mathcal{N}_{top}$ as specified in [33, Lemma 20]. Then, similarly to Lemma 6.5, we can find a rectangle solution of $\overline{\mathcal{N}}_{dir}$ that is compatible with $\mathcal{N}^*_{top}$, given that $\overline{\mathcal{N}}_{top}$ and $\overline{\mathcal{N}}_{dir}$ are satisfiable.

It is still unknown whether $\mathcal{C}_8$ is separable from DIR49. A separation theorem cannot be obtained by using a refinement mapping as for the other two subclasses. We do not regard this as a serious problem. This is because, for the purpose of backtracking, the three maximal tractable subclasses play almost the same role, and knowing one is separable is good enough to reduce the branching factor of the backtracking algorithm.

Moreover, if we confine ourselves to the less expressive cardinal direction calculus DIR9, then we have the desired separation theorems for all these subclasses. The proof is similar to that for $\widehat{\mathcal{H}}_8$ and DIR49. The interested reader may also consult Li [21] for more information.

The following example shows that, however, if $\mathcal{N}_{top}$ contains constraints not in $\widehat{\mathcal{H}}_8$, the joint network $\mathcal{N}_{top} \uplus \mathcal{N}_{dir}$ may be unsatisfiable even when both $\overline{\mathcal{N}}_{top}$ and $\overline{\mathcal{N}}_{dir}$ are satisfiable.

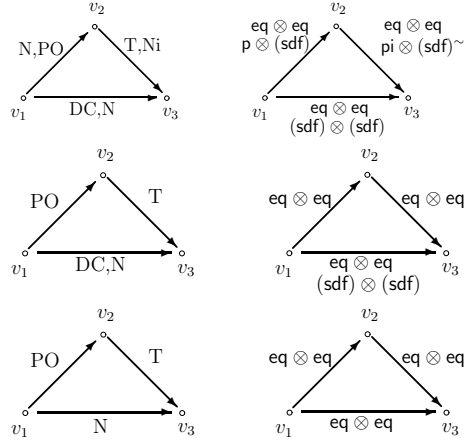

Figure 8: RCC8 network $\mathcal{N}^3_{top}$ and DIR49 network $\mathcal{N}^3_{dir}$ (first row), and their bi-closures $\overline{\mathcal{N}}^3_{top}$ and $\overline{\mathcal{N}}^3_{dir}$ (second row), and the equivalent path-consistent networks of the latter two (last row), where T, N and Ni stand for **TPP**, **NTPP** and **NTPP**$^\sim$, respectively.

**Example 7.1** (RCC8 and DIR49). Take $V = \{v_1, v_2, v_3\}$, $\mathcal{N}^3_{top} = \{v_i \theta_{ij} v_j\}^3_{i,j=1}$ and $\mathcal{N}^3_{dir} = \{v_i \delta_{ij} v_j\}^3_{i,j=1}$ are, respectively, the following two networks. (see Fig. 8)

- $\theta_{12} = \mathbf{NTPP} \cup \mathbf{PO}$, $\theta_{23} = \mathbf{TPP} \cup \mathbf{NTPP}^\sim$, $\theta_{13} = \mathbf{DC} \cup \mathbf{NTPP}$;

- $\delta_{12} = \mathsf{b} \otimes (\mathsf{sdf}) \cup \mathsf{eq} \otimes \mathsf{eq}$, $\delta_{23} = \mathsf{bi} \otimes (\mathsf{sdf})^\sim \cup \mathsf{eq} \otimes \mathsf{eq}$, $\delta_{13} = (\mathsf{sdf}) \otimes (\mathsf{sdf}) \cup \mathsf{eq} \otimes \mathsf{eq}$.

By computing $\overline{\theta}_{ij} = \theta_{ij}[\delta_{ij}]$ and $\overline{\delta}_{ij} = \delta_{ij}[\theta_{ij}]$, we obtain $\overline{\mathcal{N}}^3_{top} = \{v_i \overline{\theta}_{ij} v_j\}^3_{i,j=1}$ and $\overline{\mathcal{N}}^3_{dir} = \{v_i \overline{\delta}_{ij} v_j\}^3_{i,j=1}$ as follows.

- $\overline{\theta}_{12} = \mathbf{PO}$, $\overline{\theta}_{23} = \mathbf{TPP}$, $\overline{\theta}_{13} = \mathbf{DC} \cup \mathbf{NTPP}$;

- $\overline{\delta}_{12} = \mathsf{eq} \otimes \mathsf{eq}$, $\overline{\delta}_{23} = \mathsf{eq} \otimes \mathsf{eq}$, $\overline{\delta}_{13} = (\mathsf{sdf}) \otimes (\mathsf{sdf}) \cup \mathsf{eq} \otimes \mathsf{eq}$.

It is easy to see that $\mathcal{N}^3_{top}$ is path-consistent, and both $\overline{\mathcal{N}}^3_{top}$ and $\overline{\mathcal{N}}^3_{dir}$ are satisfiable. But $\overline{\mathcal{N}}^3_{top} \uplus \overline{\mathcal{N}}^3_{dir}$ is unsatisfiable. This is because, by applying PCA (separately) to these two networks, we refine $\overline{\theta}_{13} = \mathbf{DC} \cup \mathbf{NTPP}$ to $\mathbf{NTPP}$, and refine $\overline{\delta}_{23} = (\mathsf{sdf}) \otimes (\mathsf{sdf}) \cup \mathsf{eq} \otimes \mathsf{eq}$ to $\mathsf{eq} \otimes \mathsf{eq}$. But $\mathbf{NTPP} \cap \mathsf{eq} \otimes \mathsf{eq} = \varnothing$.

### 7.2 Beyond DIR49

So far, we have provided a complete method for deciding if a joint network of RCC8 and DIR49 constraints is satisfiable. But Figures 4 and 5 also show that we have no complete method to decide if a joint network of basic RCC8 and ERA constraints is satisfiable. In this subsection, however, we show that our results for DIR49 can also be exploited to provide approximate solutions to joint networks of RCC8 and ERA constraints.

Let $\mathcal{N}_{top} \uplus \mathcal{N}_{dir} = \{v_i \theta_{ij} v_j\}_{i,j=1}^n \uplus \{v_i \delta_{ij} v_j\}_{i,j=1}^n$ be a joint network of RCC8 and ERA constraints. Having no complete method for determining if the joint network is satisfiable, we *generalize* each ERA constraint $\delta_{ij}$ to a DIR49 constraint $\widetilde{\delta}_{ij}$, which is the smallest DIR49 relation containing $\delta_{ij}$. We call $\widetilde{\delta}_{ij}$ the *generalization* of $\delta_{ij}$ in DIR49. Write $\widetilde{\mathcal{N}}_{dir} = \{v_i \widetilde{\delta}_{ij} v_j\}_{i,j=1}^n$. We call $\widetilde{\mathcal{N}}_{dir}$ the *generalization* of $\mathcal{N}_{dir}$ in DIR49, and call $\mathcal{N}_{top} \uplus \widetilde{\mathcal{N}}_{dir}$ the generalized joint network. It is clear that a solution to $\mathcal{N}_{top} \uplus \mathcal{N}_{dir}$ is also a solution to the generalized joint network.

**Lemma 7.1.** *A joint network of RCC8 and ERA constraints is satisfiable only if its generalized joint network is.*

In other words, if the generalized joint network is not satisfiable, neither is the original one. So our separation theorems for DIR49 also provide a partial (though not complete) method for determining if a joint network of RCC8 and ERA constraints is satisfiable.

It is possible that the generalized joint network $\mathcal{N}_{top} \uplus \widetilde{\mathcal{N}}_{dir}$ is satisfiable, but $\mathcal{N}_{top} \uplus \mathcal{N}_{dir}$ itself is not. Even for this case, it is still possible to get an approximate solution to $\mathcal{N}_{dir}$.

Note that the general joint satisfaction problem (JSP) over RCC8 and ERA can be reduced to the special JSP over basic constraints by backtracking.

We only consider the case when both $\mathcal{N}_{top}$ and $\mathcal{N}_{dir}$ are basic networks. In the remainder of this subsection, we assume that

- $\mathcal{N}_{top} \uplus \mathcal{N}_{dir}$ is bi-closed and both $\mathcal{N}_{top}$ and $\mathcal{N}_{dir}$ are satisfiable;

- the generalized joint network $\mathcal{N}_{top} \uplus \widetilde{\mathcal{N}}_{dir}$ is satisfiable.

Suppose the basic ERA network $\mathcal{N}_{dir} = \{v_i \beta_{ij}^x \otimes \beta_{ij}^y v_j\}_{i,j=1}^n$. We assert that there is a solution of $\mathcal{N}_{top}$ that is *almost* a solution of $\mathcal{N}_{dir}$ in the sense that will become clear soon.

We introduce a mapping $\tau : \mathcal{B}_{int} \to \{\mathsf{b}, \mathsf{o}, \mathsf{d}, \mathsf{eq}, \mathsf{di}, \mathsf{oi}, \mathsf{bi}\}$ as follows:

$$\tau(\lambda) = \begin{cases} \mathsf{o}, & \text{if } \lambda \in \{\mathsf{m}, \mathsf{o}\}; \\ \mathsf{d}, & \text{if } \lambda \in \{\mathsf{s}, \mathsf{f}, \mathsf{d}\} \\ \mathsf{di}, & \text{if } \lambda \in \{\mathsf{si}, \mathsf{fi}, \mathsf{di}\} \\ \mathsf{oi}, & \text{if } \lambda \in \{\mathsf{mi}, \mathsf{oi}\} \\ \lambda, & \text{otherwise} \end{cases}$$

We call $\tau(\lambda)$ the $\tau$-*version* of $\lambda$. Clearly, each basic interval relation has a unique $\tau$-version.

Write $\mathcal{N}_s = \{v_i \tau(\beta_{ij}^x) \otimes \tau(\beta_{ij}^y) v_j\}_{i,j=1}^n$. Since $\mathcal{N}_{dir} = \{v_i \beta_{ij}^x \otimes \beta_{ij}^y v_j\}_{i,j=1}^n$ is satisfiable, by Lemma A.2 of Appendix B, we know $\mathcal{N}_s$ is also satisfiable. We assert that any rectangle solution $\{r_i\}_{i=1}^n$ of $\mathcal{N}_s$ is compatible with the basic RCC8 network $\mathcal{N}_{top}$.

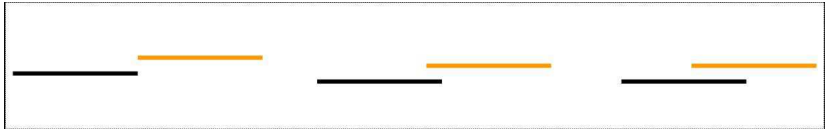

Figure 9: Illustrations of $\varepsilon$-instances of the IA relation *meets*, where the leftmost is an instance of *meets*, the middle and the right pairs are instances of *overlaps*, but the middle is more like an instance of *meets* than the right.

**Lemma 7.2.** *Suppose $\{r_i\}_{i=1}^n$ is a rectangle solution of $\mathcal{N}_s$. Then $\{r_i\}_{i=1}^n$ is compatible with $\mathcal{N}_{top}$.*

*Proof.* Since $\mathcal{N}_s = \{v_i \tau(\beta_{ij}^x) \otimes \tau(\beta_{ij}^y) v_j\}_{i,j=1}^n$ and $\tau(\beta_{ij}^x), \tau(\beta_{ij}^y) \in \{\mathsf{b}, \mathsf{o}, \mathsf{d}, \mathsf{eq}, \mathsf{di}, \mathsf{oi}, \mathsf{bi}\}$, the intersection of two rectangles $r_i$ and $r_j$ is either empty or a rectangle. It is then straightforward to show that $\{r_i\}_{i=1}^n$ is compatible with $\mathcal{N}_{top}$. For example, if $\theta_{ij} \neq \mathbf{DC}$, then by Lemma 5.2, $\mathrm{ERA}(\theta_{ij})$ contains no **MDC** relation. Since $\mathcal{N}_{top} \uplus \mathcal{N}_{dir}$ is bi-closed, we know $\delta_{ij} \subseteq \mathrm{ERA}(\theta_{ij})$. This implies that $\delta_{ij} = \beta_{ij}^x \otimes \beta_{ij}^y$ contains no **MDC** relation. By Figure 6, this is possible if and only if $\beta_{ij}^x, \beta_{ij}^y \not\in \{\mathsf{b}, \mathsf{bi}\}$. Moreover, by the definition of the $\tau$-version of an IA relation, we know $\tau(\beta_{ij}^x), \tau(\beta_{ij}^y) \not\in \{\mathsf{b}, \mathsf{bi}\}$. By Figure 6 again, $\tau(\beta_{ij}^x) \otimes \tau(\beta_{ij}^y)$ is not an **MDC** relation, i.e. $r_i \cap r_j \neq \varnothing$. Therefore, $r_i \cap r_j$ is a rectangle. □

As a corollary, we have

**Theorem 7.1.** *$\mathcal{N}_{top}$ has a solution $\{a_i\}_{i=1}^n$ which is also a solution of $\mathcal{N}_s$ and $\widetilde{\mathcal{N}}_{dir}$.*

*Proof.* Suppose $\{r_i\}_{i=1}^n$ is a rectangle solution of $\mathcal{N}_s$. By Theorem 6.2 we have a solution $\{a_i\}_{i=1}^n$ of $\mathcal{N}_{top}$ such that $\mathsf{MBR}(a_i) = r_i$ for each $i$. By the definition of the ERA relations and the assumption that $(r_i, r_j) \in \tau(\beta_{ij}^x) \otimes \tau(\beta_{ij}^y)$, we know $(a_i, a_j)$ is also an instance of the ERA relation $\tau(\beta_{ij}^x) \otimes \tau(\beta_{ij}^y)$. This shows that $\{a_i\}_{i=1}^n$ is also a solution of $\mathcal{N}_s$. Moreover, since $\mathcal{N}_s$ is a scenario of $\widetilde{\mathcal{N}}_{dir}$, we know $\{a_i\}_{i=1}^n$ is also a solution of $\widetilde{\mathcal{N}}_{dir}$. □

Although a solution of $\mathcal{N}_s$ is usually not a solution of $\mathcal{N}_{dir}$, we can find a solution of $\mathcal{N}_s$ that is *almost* a solution of $\mathcal{N}_{dir}$. The idea is to approximate a relation $\beta^x \otimes \beta^y$ by its $\tau$-version $\tau(\beta^x) \otimes \tau(\beta^y)$. Take $\mathsf{m} \otimes \mathsf{m}$ for example. Although an instance of $\mathsf{o} \otimes \mathsf{o} = \tau(\mathsf{m}) \otimes \tau(\mathsf{m})$ does not belong to $\mathsf{m} \otimes \mathsf{m}$, if $r_i \cap r_j$ is very small when compared with $r_i$ and $r_j$, then it is reasonable to say that $(r_i, r_j)$ is *almost* an instance of $\mathsf{m} \otimes \mathsf{m}$.

We formalize this idea by introducing the notion of an $\varepsilon$-instance for interval and rectangle relations (cf. Figure 9). To this end, we introduce a measure of the likeliness of an $\alpha$ instance to be a $\tau(\alpha)$ instance, where $\alpha$ is a basic IA relation.

**Definition 7.1.** For a basic IA relation $\alpha$, and an instance $(I, J)$ of $\tau(\alpha)$, we define $\chi_\alpha(I, J)$ as follows, where we assume $I = [u^-, u^+]$, $J = [v^-, v^+]$:

- If $\alpha = \mathsf{m}$, then $\tau(\alpha) = \mathsf{o}$. By $(I,J) \in \mathsf{o}$, we know $u^- < v^- < u^+ < v^+$.
  Define $\chi_{\mathsf{m}}(I,J) = (u^+ - v^-)/\min\{u^+ - u^-, v^+ - v^-\}$.

- If $\alpha = \mathsf{s}$, then $\tau(\alpha) = \mathsf{d}$. By $(I,J) \in \mathsf{d}$, we know $v^- < u^- < u^+ < v^+$.
  Define $\chi_{\mathsf{s}}(I,J) = (u^- - v^-)/(u^+ - u^-)$.

- If $\alpha = \mathsf{f}$, then $\tau(\alpha) = \mathsf{d}$. By $(I,J) \in \mathsf{d}$, we know $v^- < u^- < u^+ < v^+$.
  Define $\chi_{\mathsf{f}}(I,J) = (v^+ - u^+)/(u^+ - u^-)$.

- If $\alpha \in \{\mathsf{mi},\mathsf{si},\mathsf{fi}\}$, then define $\chi_\alpha(I,J) = \chi_{\alpha^\sim}(J,I)$, where $\alpha^\sim$ is the converse of $\alpha$.

- If $\alpha \in \{\mathsf{b},\mathsf{o},\mathsf{d},\mathsf{eq},\mathsf{di},\mathsf{oi},\mathsf{bi}\}$, then $\tau(\alpha) = \alpha$.
  Define $\chi_\alpha(I,J) = 0$.

Note that as $\chi_\alpha(I,J)$ tends to zero, then the more the instance $(I,J)$ appears to be an instance of $\alpha$. Using this measure, we next define the $\varepsilon$-instance of a basic interval relation $\alpha$.

**Definition 7.2** ($\varepsilon$-instances). For a basic interval relation $\alpha$, and an instance $(I,J)$ of $\tau(\alpha)$, we say $(I,J)$ is an $\varepsilon$-instance of $\alpha$ if $\chi_\alpha(I,J) < \varepsilon$. For a basic rectangle relation $\beta^x \otimes \beta^y$, we say an instance $(I_1 \times I_2, J_1 \times J_2)$ of $\tau(\beta^x) \otimes \tau(\beta^y)$ is an $\varepsilon$-instance of $\beta^x \otimes \beta^y$ if $(I_1, J_1)$ and $(I_2, J_2)$ are, respectively, $\varepsilon$-instances of $\beta^x$ and $\beta^y$.

The next lemma then shows that $\mathcal{N}_s = \{v_i \tau(\beta_{ij}^x) \otimes \tau(\beta_{ij}^y) v_j\}_{i,j=1}^n$ has a rectangle solution which is almost a solution of $\mathcal{N}_{dir} = \{v_i \beta_{ij}^x \otimes \beta_{ij}^y v_j\}_{i,j=1}^n$. Note that we assume $\mathcal{N}_{dir}$ is satisfiable.

**Lemma 7.3.** *For any $\varepsilon > 0$, $\mathcal{N}_s$ has a rectangle solution $\{r_i\}_{i=1}^n$ such that $(r_i, r_j)$ is an $\varepsilon$-instance of $\beta_{ij}^x \otimes \beta_{ij}^y$ for all $i,j$.*

*Proof.* We need only to prove that $\mathcal{N}_s^x = \{v_i \tau(\beta_{ij}^x) v_j\}_{i,j=1}^n$ ($\mathcal{N}_s^y = \{v_i \tau(\beta_{ij}^y) v_j\}_{i,j=1}^n$, resp.) has an interval solution $\{I_i^*\}_{i=1}^n$ ($\{J_i^*\}_{i=1}^n$, resp.) such that $(I_i^*, I_j^*)$ (($J_i^*, J_j^*$), resp.) is an $\varepsilon$-instance of $\beta_{ij}^x$ ($\beta_{ij}^y$, resp.). Take $\mathcal{N}_s^x$ as an example. Without loss of generality, we assume $\beta_{ij}^x \neq \mathsf{eq}$ for $i \neq j$.

Suppose $\{I_i = [s_{2i-1}, s_{2i}]\}_{i=1}^n$ is a solution to a basic interval network $\mathcal{N} = \{v_i \lambda_{ij}^x v_j\}_{i,j=1}^n$. We first prove that $\mathcal{N}$ has a solution $\{I_i^*\}_{i=1}^n$ that is *canonical* [43] in the following sense:

- an endpoint of each interval $I_i^*$ is an integer between $0$ and $2n-1$;

- if $k \geq 1$ is an endpoint of some interval, then $k-1$ is also an endpoint.

Clearly, each satisfiable basic interval network has a unique canonical solution.

Write $M = \{s_k\}_{k=1}^{2n}$ for the set of endpoints of all $I_i$. For $s \in M$, define its level $l(s)$ as follows:

- $l(s) = 0$ if for any $t \in M$, $s \leq t$;

- $l(s) = k+1$ if for any $t \in M$, $t < s$ only if $l(t) \leq k$.

It is straightforward to see that $l: M \to \{0, 1, \cdots, 2n-1\}$ is an order isomorphism, i.e. $l(s) \leq l(t)$ if and only if $s \leq t$. Set $I_i^* = [l(s_{2i-1}), l(s_{2i})]$. It is also straightforward to show that $\{I_i^*\}_{i=1}^n$ is the canonical solution of $\mathcal{N}$.

Now we return to $\mathcal{N}_s^x = \{v_i \tau(\beta_{ij}^x) v_j\}_{i,j=1}^n$. Suppose $\{I_i = [s_{2i-1}, s_{2i}]\}_{i=1}^n$ is a canonical solution of $\mathcal{N}_s^x$ and suppose $\{I_i' = [t_{2i-1}, t_{2i}]\}_{i=1}^n$ is a canonical solution of $\mathcal{N}_{dir}^x = \{v_i \beta_{ij}^x v_j\}_{i,j=1}^n$. Write $M = \{s_k\}_{k=1}^{2n}$ and $M' = \{t_k\}_{k=1}^{2n}$. Since $\tau(\beta_{ij}^x) \in \{\mathsf{b, o, d, oi, di, bi}\}$ for all $i \neq j$, we know $M = \{1, 2, \cdots, 2n\}$ and $s_k \neq s_p$ for any $k \neq p$.

For each $1 \leq k \leq 2n$, define $f(s_k) = t_k + \frac{s_k}{4n}\varepsilon$, where $0 < \varepsilon < 1$. Then $f: \{s_k\}_{k=1}^n \to \{f(s_k)\}_{k=1}^n$ is an order isomorphism, i.e. $f(s_k) \leq f(s_p)$ if and only if $s_k \leq s_p$. We first note that $s_k \leq s_p$ implies $t_k \leq t_p$. If $s_k \leq s_p$, then $f(s_k) = t_k + \frac{s_k}{4n}\varepsilon \leq t_p + \frac{s_p}{4n}\varepsilon = f(s_p)$. On the other hand, if $s_k > s_p$, then $t_k \geq t_p$ and $f(s_k) = t_k + \frac{s_k}{4n}\varepsilon > t_p + \frac{s_p}{4n}\varepsilon = f(s_p)$.

Set $I_i^* = [f(s_{2i-1}), f(s_{2i})]$. Then $\{I_i^*\}_{i=1}^n$ is also a solution to $\mathcal{N}_s^x$. Moreover, we can show that $\chi_\alpha(I_i^*, I_j^*) < \varepsilon$ for any $i, j$, where $\alpha = \beta_{ij}^x$. Take $\alpha = \mathsf{s}$ as an example. In this case, we have $(I_i, I_j) \in \mathsf{d}$, and $(I_i', I_j') \in \mathsf{s}$. In terms of endpoints, we have $s_{2j-1} < s_{2i-1} < s_{2i} < s_{2j}$ and $t_{2j-1} = t_{2i-1} < t_{2i} < t_{2j}$. Since $f(s_{2i-1} - f(s_{2j-1}) = t_{2i-1} + \frac{s_{2i-1}}{4n}\varepsilon - t_{2j-1} - \frac{s_{2j-1}}{4n}\varepsilon = \frac{s_{2i-1} - s_{2j-1}}{4n}\varepsilon < \varepsilon/2$, and $f(s_{2i}) - f(s_{2i-1}) = t_{2i} + \frac{s_{2i}}{4n}\varepsilon - t_{2i-1} + \frac{s_{2i-1}}{4n}\varepsilon = (t_{2i} - t_{2i-1}) + \frac{s_{2i}-s_{2i-1}}{4n}\varepsilon \geq 1$, we know $\chi_\mathsf{s}(I_i^*, I_j^*) < \varepsilon$. This means $(I_i^*, I_j^*)$ is an $\varepsilon$-instance of $\mathsf{s} = \beta_{ij}^x$. In this way, for any $i, j$, we can show $(I_i^*, I_j^*)$ is an $\varepsilon$-instance of $\beta_{ij}^x$. □

This lemma shows that $\mathcal{N}_s$ has a solution that is almost a solution of $\mathcal{N}_{dir}$. By Lemma 7.2 and Theorem 6.2, the following theorem is immediate.

**Theorem 7.2.** *Suppose $\mathcal{N}_{top} \uplus \mathcal{N}_{dir}$ is a bipath-consistent joint network of basic RCC8 and ERA constraints. If the generalized joint network $\mathcal{N}_{top} \uplus \widetilde{\mathcal{N}}_{dir}$ is satisfiable, then for any $\varepsilon > 0$, $\mathcal{N}_{top} \uplus \widetilde{\mathcal{N}}_{dir}$ has a solution $\{a_i\}_{i=1}^n$ such that $(\mathsf{MBR}(a_i), \mathsf{MBR}(a_j))$ is an $\varepsilon$-instance of $\beta_{ij}^x \otimes \beta_{ij}^y$ for all $i, j$.*

The same conclusion also holds if constraints in $\mathcal{N}_{top}$ are all taken from the maximal tractable subclass $\widehat{\mathcal{H}}_8$ of RCC8. In general, the joint satisfaction problem can be approximately determined by backtracking over $\widehat{\mathcal{H}}_8$ and $\mathcal{B}_{rec}$.

# 8 Related Work

Although most early work on qualitative spatial reasoning focused on single aspect of spatial relations, there are several works which deal with representation and reasoning about combined spatial information.

Hernández [16, 17] developed formalisms combining orientation information with topological relation or qualitative distance. Nabil et al. [27] proposed a unified representation of topological and directional relationships, based on Allen's Interval Algebra [1] and Chang's 2D string symbolical representation of pictures [3]. A similar work is also reported in Huang and Lee [18], where the authors proposed a formalism for encoding topological and directional information in a picture. We note that the direction relations defined there are exactly the same as those defined by Goyal and Egenhofer [14]. The formalism proposed in the conference version of this paper has been incorporated in the investigation of description logics with spatial operators [11].

The reasoning aspect of the combination of multiple kinds of spatial information has also been investigated by several researchers. Sharma [37] systematically studied inference problems concerning the derivation of the topological or directional relationship by given two relationships of the same or different type. An example is as follows. Suppose $a$ is a proper part of $b$ and $b$ is north of $c$. Then what kind of topological or directional relationship could hold for $a$ and $c$? Reasoning problems like this correspond to the joint satisfaction problems which involve at most three variables.

As a comparison, Sistla et al. [39, 38] considered joint satisfaction problems which involve arbitrary number of variables but are over a limited set of spatial relations. They considered connected objects in the three-dimensional space, and defined a set of part-whole relations (*disjoint, in, overlap*) and a set of three-dimensional cardinal directions (*left of, right of, above, below, in-front-of, behind*). Sistla et al. proposed a sound and complete rule-based system for determining if an arbitrary set of such constraints is satisfiable as connected objects in three-dimensional space, where several constraints concerning the same pair of variables may appear at the same time. As for two-dimensional space, they showed that the rule-based system is incomplete for connected plane regions. But it is straightforward to show that the rule-based system is complete when instantiations are taken from the universe of bounded (connected or disconnected) plane regions.

Write $\mathcal{T}$ for the set of part-whole relations *disjoint, in, overlap*, and write $\mathcal{D}$ for the set of cardinal directions *left of, right of, above, below*. Clearly, $\mathcal{T}$ is a subset of RCC5 (hence of RCC8), and $\mathcal{D}$ is a subset of DIR9 (hence of ERA). Write $\widehat{\mathcal{T}}$ ($\widehat{\mathcal{D}}$, resp.) for the smallest subclass of RCC8 (ERA, resp.) containing $\mathcal{T}$ ($\mathcal{D}$, resp.) which is closed under converse and intersection. Then, the contribution of Sistla et al. can be rephrased as providing a complete method for determining the JSPs over $\widehat{\mathcal{T}}$ and $\widehat{\mathcal{D}}$.

Compared with $\widehat{\mathcal{H}}_8$ and DIR49, this constraint language is very small. More important, the topological part ($\widehat{\mathcal{T}}$) makes no further topological distinction between, e.g. *tangential proper part* (**TPP**) and *non-tangential proper part* (**NTPP**); and the directional part ($\widehat{\mathcal{D}}$) does not support negation and disjunction of constraints, i.e. constraints such as *not left of* and *either right of or above* are not allowed in their constraint language.

Another attempt to combining topological and directional information was reported in [19], where the author introduced a hybrid calculus that combines DIR9 with RCC5. A preliminary result was obtained, which asserts that the satisfaction problem of basic networks in the hybrid calculus can be decided in polynomial time. This is equivalent to say that the joint satisfaction problem of basic RCC5 and DIR9 networks can be decided in polynomial time. The work reported in the current paper is more general.

The BIPATH-CONSISTENCY algorithm was first introduced by Gerevini and Renz [12], where they discussed the combination of topological and relative size information, and proved that BIPATH-CONSISTENCY is complete for the JSPs over any maximal tractable subclass of RCC8 and the qualitative size calculus QS. In this paper we gave a characterization of bipath-consistency in terms of bi-closure and path-consistency, and hence generalized the algorithm to cope with two arbitrary qualitative calculi.

*Remark* 8.1. Recently, Wölfl and Westphal [42] also investigated the combination of binary qualitative constraint calculi in general, where they empirically compared the (tight combination) approach that develops a new hybrid calculus with the (loose combination) approach of Gereveni and Renz [12]. Note the latter approach is also known the joint satisfaction problem in this paper. Our research in this paper is mainly concerned with the loose combination of topological and directional constraints, while the early work of Li [19] provided an example of a tight combination.

# 9 Conclusion and Future Work

In this paper, we have investigated computational complexity of reasoning with the combination of a topological relation calculus (RCC8 Algebra) and a directional relation calculus (Extended Rectangle Algebra ERA). We first showed by examples that BIPATH-CONSISTENCY is incomplete for solving the JSP over even basic RCC8 and ERA constraints topological constraints from directional constraints as one key problem for solving the joint satisfaction problem over RCC8 and ERA, and then proved that for two maximal tractable subclasses of RCC8 ($\widehat{\mathcal{H}}_8$ or $\mathcal{Q}_8$) and a subalgebra of ERA (DIR49) BIPATH-CONSISTENCY separates topological constraints in polynomial time from directional constraints. Therefore, the joint satisfaction problem of a network of constraints over $\widehat{\mathcal{H}}_8$ (or $\mathcal{Q}_8$) and DIR49 can be reduced in polynomial time to two simple satisfaction problems in RCC8 and ERA.

The fact that $\widehat{\mathcal{H}}_8$ (or $\mathcal{Q}_8$) is separable from DIR49 implicitly suggests that the interaction between RCC8 and DIR49 is weak. Naturally, if the interaction between two calculi is very strong, then it will be hopeless to get a clear separation between them. Moreover, just like the interaction between the qualitative size calculus and RCC8 [12], DIR49 relations interact with RCC5 more than RCC8.[3] This is because we often ignore the boundary of regions in DIR49.

For our purposes this weakness is a not serious problem. Particularly, for RCC8 and ERA, we take the view that "*topology matters, metric refines* [9]." For a satisfiable joint network of basic RCC8 and ERA constraints, we can always find an instantiation that satisfies all topological constraints and almost satisfies all directional constraints. We believe this is good enough for most practical applications.

Although BIPATH-CONSISTENCY is incomplete for the JSP of RCC8 and ERA, this does not mean that reasoning with RCC8 and ERA is undecidable.

---

[3]One exception is the rule that $a\mathbf{NTPP}b$ implies $(\mathsf{MBR}(a), \mathsf{MBR}(b)) \in \mathsf{d} \otimes \mathsf{d}$.

Recently, Liu et al. [26] proved that the JSP for *basic* RCC8 constraints and *basic* ERA constraints is still tractable. More work is needed in this direction to discover larger tractable subclasses.

Another possible weakness of this paper lies in the use of rectangle relations to approximate direction between two arbitrarily shaped regions. This is over simplistic for many real-world applications. The cardinal direction calculus (CDC) of Goyal and Egenhofer [14] is a very expressive spatial language for directions, and its computational complexity has just been investigated very recently [40, 43]. For basic RCC8 constraints and basic CDC constraints, Liu et al. [26] proved that the joint satisfaction problem is already NP-Complete. Therefore, approximative but efficient methods similar to the one proposed in Section 7.2 of this paper will be very useful to cope with combined RCC8 and CDC constraints.

Since BIPATH-CONSISTENCY separates (to a certain extent) topological information from both directional (DIR49) and qualitative size information, it is natural to extend the results obtained here and that in [12] to cope with the combination of relations in the three calculi RCC8, ERA, and QS. We remark that such a combination is straightforward since there is no direct interaction between ERA and QS constraints.

## Acknowledgement

We gratefully acknowledge the Royal Society for the financial support of a short visit from the first author to the second. The work of Sanjiang Li was also partially supported by NSFC (60673105, 60621062). The work of Tony Cohn was also partially supported by EP/D061334/1.

## A  Proof of Theorem 7.1

Recall $\tau : \mathcal{B}_{int} \to \{\mathsf{b}, \mathsf{o}, \mathsf{d}, \mathsf{eq}, \mathsf{di}, \mathsf{oi}, \mathsf{bi}\}$ is defined as follows:

$$\widehat{\mathsf{b}} = \mathsf{b},\ \widehat{\mathsf{m}} = \widehat{\mathsf{o}} = \mathsf{o},\ \widehat{\mathsf{s}} = \widehat{\mathsf{d}} = \widehat{\mathsf{f}} = \mathsf{d},\ \widehat{\mathsf{eq}} = \mathsf{eq},\ \widehat{\mathsf{eq}} = \mathsf{eq},\ \widehat{\mathsf{si}} = \widehat{\mathsf{di}} = \widehat{\mathsf{fi}} = \mathsf{di}, \qquad (23)$$

where for convenience we write $\widehat{\beta}$ for $\tau(\beta)$, the $\tau$-version of $\beta$. For a basic IA network $\mathcal{N} = \{x_i \lambda x_j\}_{1 \leq i,j \leq n}$, write $\widehat{\mathcal{N}}$ for the basic IA network $\{x_i \widehat{\lambda} x_j\}_{i,j=1}^n$, called the $\tau$-version of $\mathcal{N}$. Then we have the following interesting result.

**Lemma A.1.** *A basic IA network $\mathcal{N}$ is satisfiable only if its $\tau$-version $\widehat{\mathcal{N}}$ is.*

*Proof.* If $\mathcal{N}$ involves only three variables (a triangle), the proof is straightforward. So each sub-network of $\widehat{\mathcal{N}}$ involving three variables are satisfiable. In general, recall that a basic IA network is satisfiable if and only if it is path-consistent. This implies that each triangle in $\widehat{\mathcal{N}}$ is path-consistent. By definition of path-consistency, the whole network is path-consistent, hence satisfiable. □

For a basic rectangle relation $\delta = \alpha \otimes \beta$ we call $\widehat{\alpha} \otimes \widehat{\beta}$ the $\tau$-version of $\delta$, denoted by $\widehat{\delta}$. For example, the $\tau$-version of $\mathsf{eq} \otimes \mathsf{s}$ is $\mathsf{eq} \otimes \mathsf{d}$.

**Lemma A.2.** *A basic RA network is consistent only if its $\tau$-version is.*

The definition of $\tau$-version can be extended to non-basic relations in a natural way. Let $\alpha$ be an IA or RA (non-basic) relation, the $\tau$-version of $\alpha$, denoted by $\widehat{\alpha}$, is defined as

$$\widehat{\alpha} = \bigcup \{\widehat{\beta} : \beta \text{ is a basic relation and } \beta \subseteq \alpha\}.$$

The $\tau$-version of an IA or RA network is defined similarly.

**Lemma A.3.** *An IA or RA network is satisfiable only if its $\tau$-version is.*

For an IA or RA relation $\alpha$, we say $\alpha$ is $\tau$-*closed* if it contains its $\tau$-version, i.e. $\widehat{\alpha} \subseteq \alpha$. Similarly, an IA or RA network is $\tau$-*closed* if all its constraints are $\tau$-closed.

The following lemmas are easy to check.

**Lemma A.4.** *For an RCC8 relation $\theta$, $\mathrm{ERA}(\theta)$ is $\tau$-closed, where $\mathrm{ERA}(\theta)$ is the smallest ERA relation which contains $\theta$.*

**Lemma A.5.** *Each DIR49 relation is $\tau$-closed.*

Since the intersection of two $\tau$-closed relations is also $\tau$-closed, by the above lemmas we have

**Lemma A.6.** *For an RCC8 relation $\theta$ and a DIR49 relation $\delta$, $\delta[\theta]$ is $\tau$-closed, where $\delta[\theta] = \delta \cap \mathrm{ERA}(\theta)$.*

The next theorem follows directly from Lemma A.3.

**Theorem A.1.** *Let $\mathcal{N} = \{v_i \delta_{ij} v_j\}_{i,j=1}^n$ be a $\tau$-closed RA network. If $\mathcal{N}$ is satisfiable, then it has a satisfiable scenario $\mathcal{N}' = \{v_i \delta'_{ij} v_j\}_{i,j=1}^n$ such that each $\delta'_{ij}$ has the form $\lambda_{ij}^x \otimes \lambda_{ij}^y$, where $\lambda_{ij}^x, \lambda_{ij}^y \in \{\mathsf{b}, \mathsf{o}, \mathsf{d}, \mathsf{eq}, \mathsf{di}, \mathsf{oi}, \mathsf{bi}\}$.*

*Proof.* By Lemma A.3, the $\tau$-version of $\mathcal{N}$ is also satisfiable. This implies it has a satisfiable scenario $\mathcal{N}'$ which satisfies the above condition. □

Recall that an RA network is satisfiable if and only if its corresponding ERA network is (see Lemma 3.1). As a corollary of Theorem A.1 and Lemma A.6, we have

**Theorem A.2** (Theorem 7.1). *Let $\mathcal{N}_{top}$ be a path-consistent RCC8 network, and let $\mathcal{N}_{dir}$ be a DIR49 network. Suppose $\overline{\mathcal{N}}_{dir}$ is satisfiable. Then $\overline{\mathcal{N}}_{dir}$ has a satisfiable scenario $\mathcal{N}'_{dir} = \{v_i \delta'_{ij} v_j\}_{i,j=1}^n$ such that each $\delta'_{ij}$ has the form $\lambda_{ij}^x \otimes \lambda_{ij}^y$, where $\lambda_{ij}^x, \lambda_{ij}^y \in \{\mathsf{b}, \mathsf{o}, \mathsf{d}, \mathsf{eq}, \mathsf{di}, \mathsf{oi}, \mathsf{bi}\}$.*

*Proof.* Because $\overline{\mathcal{N}}_{dir}$ is $\tau$-closed, the conclusion follows directly from Theorem A.1. □

# B  Proof of Theorem 7.2

**Theorem B.1** (Theorem 7.2). *Let $\mathcal{N}_{top} = \{v_i \theta_{ij} v_j\}_{i,j=1}^n$ be a satisfiable RCC8 basic network. Suppose $\{r_i\}_{i=1}^n$ is a collection of rectangles that are compatible with $\mathcal{N}_{top}$. Then we have a solution $\{a_i^*\}_{i=1}^n$ of $\mathcal{N}_{top}$ such that each $a_i^*$ is a bounded region and $\mathsf{MBR}(a_i^*) = r_i$ for any $1 \leq i \leq n$.*

*Proof.* The proof is similar to that given for RCC8 (cf. [32, 20, 22]). First, we define $l(i)$, the ntpp-level of $v_i$, inductively as follows:

- $l(i) = 1$ if there is no $j$ such that $\theta_{ji} = \mathbf{NTPP}$;

- $l(i) = k+1$ if there is a variable $v_j$ such that (a) $l(j) = k$ and $\theta_{ji} = \mathbf{NTPP}$; and (b) $\theta_{mi} = \mathbf{NTPP}$ implies $l(m) \leq k$ for any variable $v_m$.

For each rectangle $r_i$, we write $e_{il}$ ($E_{il}$, resp.) ($l = 1, 2, 3, 4$) for the four edge (corner points, resp.) of $r_i$. Moreover, for each edge $e_{il}$, we choose $n$ points $P_{il}^j$ ($1 \leq j \leq n$) such that

- if $i \neq i'$ or $j \neq j'$ or $l \neq l'$, then $P_{il}^j$ and $P_{i'l'}^{j'}$ are distinct;

- no $P_{il}^j$ is a corner point of any rectangle $r_k$.

Furthermore, for $i \neq j$, if $\theta_{ij}$ is **EC** or **PO**, we choose two new points $Q_{ij}$ and $Q_{ji}$ in the interior of $r_i \cap r_j$ such that $Q_{ij}$ and $Q_{ji}$ are not in any edge of any rectangle $r_k$. Set $N$ to be the set of all these points $E_{il}$, $P_{il}^j$, $Q_{ij}$, and set $\delta_1 > 0$ to be the smallest distance between two points in $N$.

For a point $P$ in $N$, and an edge $e_{il}$ of a rectangle $r_i$, if $P$ is not in $e_{il}$, then $d(P, e_{il}) \equiv \min\{d(P, P') : P' \in e_{il}\}$, the distance from $P$ to $e_{il}$, is nonzero. Therefore the distance from any point $P$ in $N$ to any edge $e_{il}$ with $P \notin e_{il}$ is bigger than a positive real number, say $\delta_2$.

Choose $\delta > 0$ smaller than both $\delta_1$ and $\delta_2$. For each point $P$ in $N$, construct a system of concentric disks $\{p^{(1)}, \cdots, p^{(n)}\}$ as in Figure 10, where $p^{(i)}$ is a disk centered at $P$ with radius $r_i$ such that $0 < r_1 < r_2 < \cdots < r_n < \delta/4$. If $\theta_{ij} = \mathbf{EC}$ and $P = Q_{ij}$, then write $q_{ij}^-$ and $q_{ij}^+$ for the left and right halves of the disk $q_{ij}^{(1)}$.

Now we construct $n$ bounded regions $\{a_i^*\}_{i=1}^n$ as follows.

- $a_i = r_i \cap \bigcup_{k=1}^4 p_{il}^{(1)}$;

- $a_i' = a_i \cup \bigcup \{q_{ij}^{(-)} \cup q_{ji}^{(+)} : \theta_{ij} = \mathbf{EC}\} \cup \bigcup \{q_{ij}^{(1)} \cup q_{ji}^{(1)} : \theta_{ij} = \mathbf{PO}\}$;

- $a_i'' = a_i' \cup \{a_k' : \theta_{ki}$ is **TPP** or **NTPP**$\}$;

- $a_i^* = a_i'' \cup \bigcup \{p^{(l(i))} : P \in N$ and $(\exists j)[\theta_{ji} = \mathbf{NTPP}$ and $p^{(1)} \cap a_j'' \neq \varnothing]\}$.

Then $\{a_i^*\}_{i=1}^n$ is a solution of $\mathcal{N}_{top}$. Moreover, we have $r_i = \mathsf{MBR}(a_i^*)$.  □

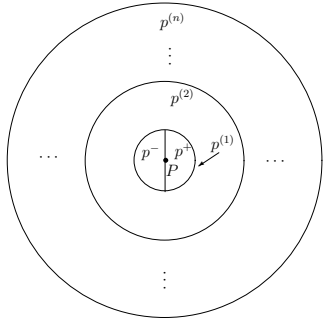

Figure 10: An illustration of the **NTPP**-chain centered at $P$.

## C  Proof of Lemma 7.3

**Lemma C.1** (Lemma 7.3). *For an RCC8 relation $\theta \in \widehat{\mathcal{H}}_8$ and an ERA relation $\delta$, if $\theta[\delta] \neq \varnothing$ and $\delta[\theta] \neq \varnothing$, then $\hbar(\theta) = \hbar(\theta[\delta])$.*

*Proof.* We prove this case by case.

- If $\mathbf{DC} \subseteq \theta$, we assert that $\mathbf{DC}$ is contained in $\mathrm{RCC}(\delta)$, hence in $\theta[\delta] = \theta \cap \mathrm{RCC}(\delta)$. This is because, by Lemma 5.3, $\mathbf{DC}$ is contained in $\mathrm{RCC}(\delta')$ for any basic ERA relation $\delta'$. By definition of $\hbar$ we know $\hbar(\theta[\delta]) = \mathbf{DC}$.

- If $\mathbf{DC} \cap \theta = \varnothing$ but $\mathbf{EC} \subseteq \theta$, we assert that $\mathbf{EC}$ is contained in $\mathrm{RCC}(\delta)$, hence contained in $\theta[\delta]$. This is because, by Lemma 5.3, $\mathbf{EC}$ is contained in each $\mathrm{RCC}(\delta')$ for any basic ERA relation $\delta'$ that is not an $\mathbf{MDC}$ relation. Moreover, since $\theta[\delta] = \theta \cap \mathrm{RCC}(\delta)$ is nonempty, $\mathrm{RCC}(\delta) \not\subseteq \mathbf{DC}$. This implies that $\delta$ contains a non-$\mathbf{MDC}$ basic ERA relation. Therefore $\mathbf{EC} \subseteq \mathrm{RCC}(\delta)$.

  By definition of $\hbar$ we know $\hbar(\theta[\delta]) = \mathbf{EC}$.

- If $(\mathbf{DC} \cup \mathbf{EC}) \cap \theta = \varnothing$ but $\mathbf{PO} \subseteq \theta$, we assert that $\mathbf{PO}$ is contained in $\mathrm{RCC}(\delta)$, hence contained in $\theta[\delta]$. This is because, by Lemma 5.3, $\mathbf{PO}$ is contained in each $\mathrm{RCC}(\delta')$ for any basic ERA relation $\delta'$ that is neither an $\mathbf{MDC}$ nor an $\mathbf{MEC}$ relation. Moreover, since $\theta[\delta] = \theta \cap \mathrm{RCC}(\delta)$ is nonempty, $\mathrm{RCC}(\delta) \not\subseteq \mathbf{DC} \cup \mathbf{EC}$. This implies that $\delta$ contains a basic ERA relation that is neither $\mathbf{MDC}$ nor $\mathbf{MEC}$. Therefore $\mathbf{PO} \subseteq \mathrm{RCC}(\delta)$.

  By definition of $\hbar$ we know $\hbar(\theta[\delta]) = \mathbf{PO}$.

- If $(\mathbf{DC} \cup \mathbf{EC} \cup \mathbf{PO}) \cap \theta = \varnothing$ but $\mathbf{TPP} \subseteq \theta$, we assert that $\mathbf{TPP}$ is contained in $\mathrm{RCC}(\delta)$, hence contained in $\theta[\delta]$. This is because for a basic ERA relation $\delta'$, by Lemma 5.3, $\mathbf{TPP}$ is contained in $\mathrm{RCC}(\delta')$ if and only if $\delta'$ is an $\mathbf{MTPP}$ or $\mathbf{MNTPP}$ or $\mathbf{MEQ}$ relation. Since $\theta$ is in $\widehat{\mathcal{H}}_8$, it must be contained in $\mathbf{P}$. Furthermore, since $\theta[\delta] = \theta \cap \mathrm{RCC}(\delta)$ is nonempty, $\mathbf{P} \cap \mathrm{RCC}(\delta) \neq \varnothing$. This is possible only if $\delta$ contains a basic

ERA relation that is either **MTPP** or **MNTPP** or **MEQ**. In each case, we have **TPP** $\subseteq \text{RCC}(\delta)$.

By definition of $\hbar$ we know $\hbar(\theta[\delta]) = \textbf{TPP}$.

- The case when $(\textbf{DC} \cup \textbf{EC} \cup \textbf{PO}) \cap \theta = \varnothing$ but $\textbf{TPP}^\sim \subseteq \theta$ is similar.

- For all the other cases, we know $\theta$ must be a basic relation. Since $\theta \supseteq \theta[\delta] \neq \varnothing$, we know $\theta[\delta] = \theta$. That is, we also have $\hbar(\theta[\delta]) = \hbar(\theta)$ in this case.

This ends the proof. $\square$

# References

[1] J.F. Allen. Maintaining knowledge about temporal intervals. *Communications of the ACM*, 26:832–843, 1983.

[2] P. Balbiani, J.-F. Condotta, and L. Fariñas del Cerro. A new tractable subclass of the rectangle algebra. In D. Dean, editor, *Proceedings of the Sixteenth International Joint Conference on Artificial Intelligence (IJCAI-99)*, pages 442–447. Morgan Kaufmann, 1999.

[3] S.K. Chang, Q.Y. Shi, and C.W. Yan. Iconic indexing by 2-d strings. *IEEE Transactions on Pattern Analysis and Machine Intelligence*, 9(3):413–428, 1987.

[4] A.G. Cohn and S.M. Hazarika. Qualitative spatial representation and reasoning: An overview. *Fundamenta Informaticae*, 46(1-2):1–29, 2001.

[5] A.G. Cohn and J. Renz. Qualitative spatial reasoning. In F. van Harmelen, V. Lifschitz, and B. Porter, editors, *Handbook of Knowledge Representation*. Elsevier, 2007.

[6] R. Dechter. *Constraint processing*. Morgan Kaufmann Publishers, San Francisco, CA, 2003.

[7] I. Düntsch, H. Wang, and S. McCloskey. A relation-algebraic approach to the Region Connection Calculus. *Theoretical Computer Science*, 255:63–83, 2001.

[8] M.J. Egenhofer. A formal definition of binary topological relationships. In *Proceedings of the Third International Conference on Foundations of Data Organization and Algorithms*, Paris, 1989.

[9] M.J. Egenhofer and D.M. Mark. Naive geography. In A.U. Frank and W. Kuhn, editors, *Proceedings of the Second International Conference on Spatial Information Theory (COSIT-95)*, pages 1–15, 1995.

[10] C. Freksa. Temporal reasoning based on semi-intervals. *Artif. Intell.*, 54(1):199–227, 1992.

[11] N. Gabrielli. *Investigation of the Tradeoff between Expressiveness and Complexity in Description Logics with Spatial Operators*. Ph.D thesis, Università degli Studi di Verona, May 2009.

[12] A. Gerevini and J. Renz. Combining topological and size information for spatial reasoning. *Artificial Intelligence*, 137(1):1–42, 2002.

[13] M.C. Golumbic and R. Shamir. Complexity and algorithms for reasoning about time: a graph-theoretic approach. *Journal of the ACM*, 40(5):1108–1133, 1993.

[14] R. Goyal and M.J. Egenhofer. Similarity of cardinal directions. In C.S. Jensen, M. Schneider, B. Seeger, and V.J. Tsotras, editors, *Proceedings of the 7th International Symposium on Advances in Spatial and Temporal Databases (SSTD-01)*, pages 36–58. Springer, 2001.

[15] H.W. Guesgen. Spatial reasoning based on allen's temporal logic. Technical report, International Computer Science Institute, 1989.

[16] D. Hernández. *Qualitative Representation of Spatial Knowledge*, volume 804 of *Lecture Notes in Computer Science*. Springer, 1994.

[17] D. Hernández, E. Clementini, and P. Di Felice. Qualitative distances. In A.U. Frank and W. Kuhn, editors, *Proceedings of the Second International Conference on Spatial Information Theory (COSIT-95)*, pages 45–57. Springer, 1995.

[18] P.W. Huang and C.H. Lee. Image database design based on 9D-SPA representation for spatial relations. *IEEE Transactions on Knowledge and Data Engineering*, 16(12):1486–1496, 2004.

[19] S. Li. Combining topological and directional information: First results. In J. Lang, F. Lin, and J. Wang, editors, *Proceedings of the First International Conference on Knowledge Science, Engineering and Management (KSEM-06)*, pages 252–264. Springer, 2006.

[20] S. Li. On topological consistency and realization. *Constraints*, 11(1):31–51, 2006.

[21] S. Li. Combining topological and directional information for spatial reasoning. In M. Veloso, editor, *Proceedings of the 20th International Joint Conference on Artificial Intelligence (IJCAI-07)*, pages 435–440. AAAI, 2007.

[22] S. Li and H. Wang. RCC8 binary constraint network can be consistently extended. *Artificial Intelligence*, 170(1):1–18, 2006.

[23] S. Li and M. Ying. Extensionality of the RCC8 composition table. *Fundamenta Informaticae*, 55(3):363–385, 2003.

[24] S. Li and M. Ying. Region Connection Calculus: Its models and composition table. *Artificial Intelligence*, 145(1-2):121–146, 2003.

[25] G. Ligozat and J. Renz. What is a qualitative calculus? A general framework. In C. Zhang, H. Guesgen, and W.-K. Yeap, editors, *Proceedings of the 8th Pacific Rim Trends in Artificial Intelligence (PRICAI-04)*, pages 53–64. Springer, 2004.

[26] W. Liu, S. Li, and J. Renz. Combining RCC-8 with qualitative direction calculi: Algorithms and complexity. In C. Boutilier, editor, *Proceedings of the Twenty-first International Joint Conference on Artificial Intelligence (IJCAI-09)*, pages 854–859, 2009.

[27] M. Nabil, J. Shepherd, and A. Ngu. 2d projection interval relationships: A symbolic representation of spatial relationships. In M.J. Egenhofer and J.R. Herring, editors, *Proceedings of the Fourth International Symposium on Advances in Spatial Databases (SSD-95)*, pages 292–309. Springer, 1995.

[28] B. Nebel and H.-J. Bürckert. Reasoning about temporal relations: A maximal tractable subclass of Allen's interval algebra. *Journal of the ACM*, 42(1):43–66, 1995.

[29] R. Maddux P.B. Ladkin. On binary constraint problems. *Journal of the ACM*, 41(3):435–469, 1994.

[30] D.J. Peuquet and C.-X. Zhan. An algorithm to determine the directional relationship between arbitrarily-shaped polygons in the plane. *Pattern Recognition*, 20(1):65–74, 1987.

[31] D.A. Randell, Z. Cui, and A.G. Cohn. A spatial logic based on regions and connection. In *Proceedings of the Third International Conference on Principles of Knowledge Representation and Reasoning (KR-92)*, pages 165–176, 1992.

[32] J. Renz. A canonical model of the region connection calculus. In *Proceedings of the 6th International Conference on Principles of Knowledge Representation and Reasoning (KR-98)*, pages 330–341. Morgan Kaufmann, 1998.

[33] J. Renz. Maximal tractable fragments of the Region Connection Calculus: A complete analysis. In D. Dean, editor, *Proceedings of the Sixteenth International Joint Conference on Artificial Intelligence (IJCAI-99)*, pages 448–454. Morgan Kaufmann, 1999.

[34] J. Renz. *Qualitative spatial reasoning with topological information*, volume 2293 of *Lecture Notes in Artificial Intelligence*. Springer-Verlag, Berlin, Germany, 2002.

[35] J. Renz and B. Nebel. On the complexity of qualitative spatial reasoning: A maximal tractable fragment of the Region Connection Calculus. *Artificial Intelligence*, 108:69–123, 1999.

[36] R. Renz and F. Schmid. Customizing qualitative spatial and temporal calculi. In Mehmet A. Orgun and John Thornton, editors, *Australian Conference on Artificial Intelligence*, volume 4830 of *Lecture Notes in Computer Science*, pages 293–304. Springer, 2007.

[37] J. Sharma. *Integrated Spatial Reasoning in Geographic Information Systems: Combining Topology and Direction*. Ph.D thesis, University of Maine, May 1996.

[38] A.P. Sistla and C.T. Yu. Reasoning about qualitative spatial relationships. *Journal of Automated Reasoning*, 25(4):291–328, 2000.

[39] A.P. Sistla, C.T. Yu, and R. Haddad. Reasoning about spatial relationships in picture retrieval systems. In J.B. Bocca, M. Jarke, and C. Zaniolo, editors, *Proceedings of 20th International Conference on Very Large Data Bases (VLDB-94)*, pages 570–581. Morgan Kaufmann, 1994.

[40] S. Skiadopoulos and M. Koubarakis. On the consistency of cardinal direction constraints. *Artificial Intelligence*, 163(1):91–135, 2005.

[41] S. Skiadopoulos, N. Sarkas, T.K. Sellis, and M. Koubarakis. A family of directional relation models for extended objects. *IEEE Trans. Knowl. Data Eng.*, 19(8):1116–1130, 2007.

[42] S. Wölfl and M. Westphal. On combinations of binary qualitative constraint calculi. In C. Boutilier, editor, *Proceedings of the Twenty-first International Joint Conference on Artificial Intelligence (IJCAI-09)*, pages 967–972, 2009.

[43] X. Zhang, W. Liu, S. Li, and M. Ying. Reasoning with cardinal directions: An efficient algorithm. In D. Fox and C. Gomes, editors, *Proceedings of the Twenty-Third AAAI Conference on Artificial Intelligence (AAAI-08)*, 2008.